%% file: main_neurips_2024.tex
\documentclass{article}

    \PassOptionsToPackage{numbers, compress}{natbib}

\usepackage[preprint]{neurips_2024}

\usepackage[utf8]{inputenc} 
\usepackage[T1]{fontenc}    
\usepackage{hyperref}       
\usepackage{url}            
\usepackage{booktabs}       
\usepackage{amsfonts}       
\usepackage{nicefrac}       
\usepackage{microtype}      
\usepackage{adjustbox}
\usepackage{soul}
\usepackage[table]{xcolor} 

\usepackage{amsmath}
\usepackage{amssymb}
\usepackage{mathtools}
\usepackage{amsthm}
\usepackage{pdfpages}
\usepackage{changepage} 
\usepackage{wrapfig}
\usepackage{graphicx}
\usepackage{subcaption}
\usepackage{enumitem}
\usepackage{float}
\usepackage{pdflscape}
\usepackage{rotating}

\usepackage{graphicx}

\title{On the Scalability of GNNs for Molecular Graphs}

\author{%
  Maciej Sypetkowski\thanks{Equal contribution; order determined alphabetically.} \\
  Valence Labs, Montreal \\
  \texttt{maciej@valencelabs.com } \\
  \And
  Frederik Wenkel\footnotemark[1] \\
  Valence Labs, Montreal \\
  Université de Montréal, Mila Quebec \\
  \texttt{frederik@valencelabs.com} \\
  \AND
  Farimah Poursafaei \\
  Valence Labs, Montreal \\
  McGill University, Mila Quebec \\
  \And
  Nia Dickson \\
  NVIDIA Corporation \\
  \And
  Karush Suri \\
  Valence Labs, Montreal \\
  \And
  Philip Fradkin \\
  Valence Labs, Montreal \\
  University of Toronto, Vector Institute \\
  \And
  Dominique Beaini \\
  Valence Labs, Montreal \\
  Université de Montréal, Mila Quebec \\
}

\definecolor{dark2green}{RGB}{27, 158, 119}
\definecolor{dark2blue}{RGB}{27, 48, 158}
\definecolor{dark2orange}{RGB}{237, 127, 49}

\input{macros}

\begin{document}

\maketitle

\begin{abstract}

\input{sections/000_abstract}

\end{abstract}

\input{sections/001_introduction}

\input{sections/002_related_work}

\input{sections/003_methodology}

\input{sections/004_scaling_laws}

\input{sections/006_discussion}





\bibliography{refs}
\bibliographystyle{abbrvnat}



\input{sections/009_Appendix}





\end{document}

%% file: macros.tex
\NewDocumentCommand{\codeword}{v}{%
\texttt{\textcolor{red}{#1}}%
}

\NewDocumentCommand{\tocheck}{v}{%
\textcolor{red}{#1}%
}

%% file: sections/000_abstract.tex

Scaling deep learning models has been at the heart of recent revolutions in language modelling and image generation. Practitioners have observed a strong relationship between model size, dataset size, and performance. However, structure-based architectures such as Graph Neural Networks (GNNs) are yet to show the benefits of scale mainly due to the lower efficiency of sparse operations, large data requirements, and lack of clarity about the effectiveness of various architectures. 
We address this drawback of GNNs by studying their scaling behavior. Specifically, we analyze message-passing networks, graph Transformers, and hybrid architectures on the largest public collection of 2D molecular graphs. 
For the first time, we observe that GNNs benefit tremendously from the increasing scale of depth, width, number of molecules, number of labels, and the diversity in the pretraining datasets.
We further demonstrate strong finetuning scaling behavior on 38 highly competitive downstream tasks, outclassing previous large models.
This gives rise to \emph{MolGPS}, a new graph foundation model that allows to navigate the chemical space, outperforming the previous state-of-the-arts on 26 out the 38 downstream tasks.
We hope that our work paves the way for an era where foundational GNNs drive pharmaceutical drug discovery.

%% file: sections/001_introduction.tex

\section{Introduction}
\label{sec:introduction}


\begin{wrapfigure}{R}{0.6\columnwidth}
\vspace{-10pt}
\centering
    \includegraphics[width=\linewidth]{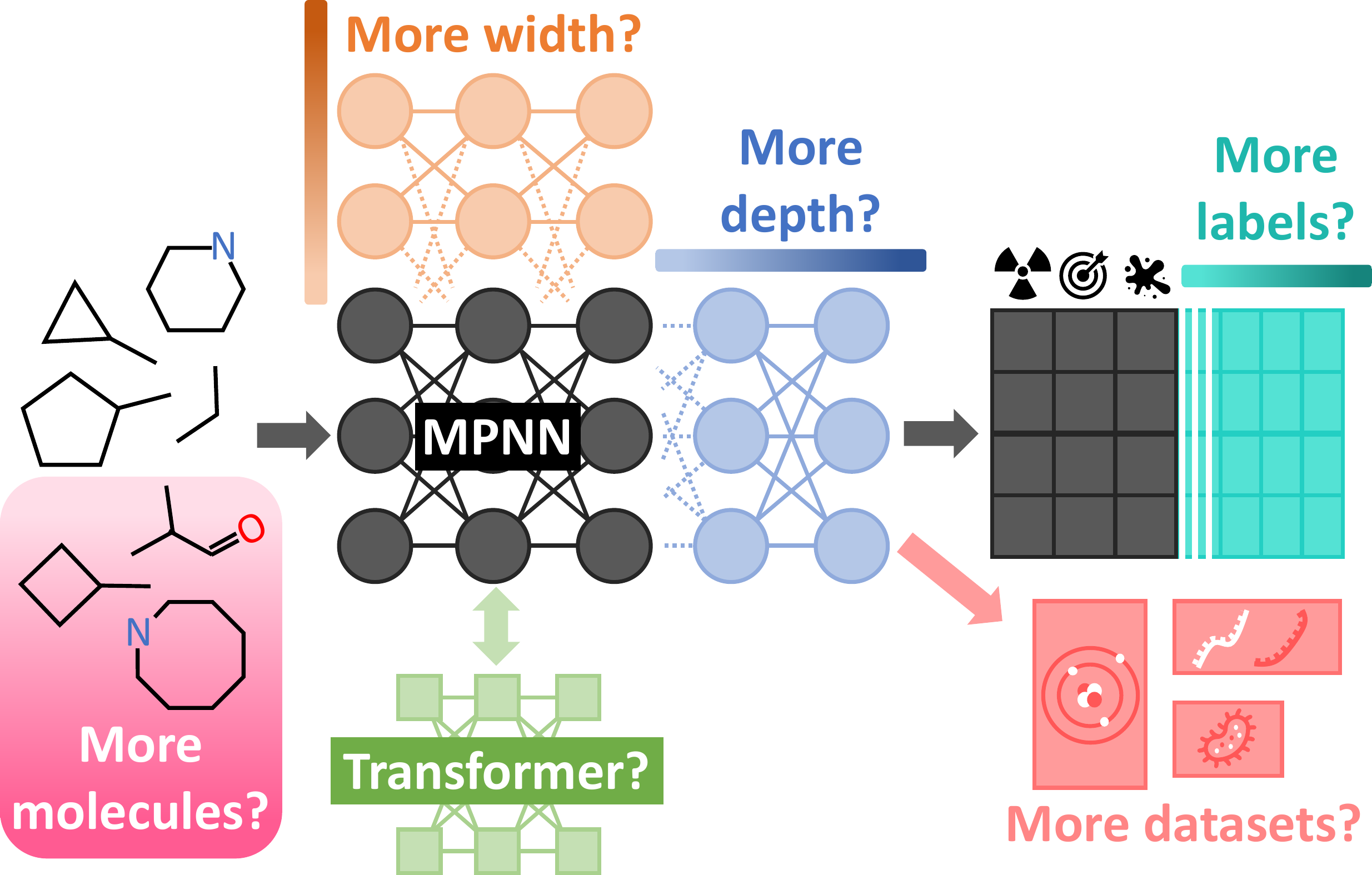}
    \caption{
     Summary of our GNN scaling hypotheses studied in the present work. The baseline model is presented in \textbf{\color{gray}{dark grey}}, followed by different scaling hypotheses illustrated in lighter colors. We analyze the scaling behavior of message-passing networks, graph Transformers and hybrid architectures with respect to the increasing scale of width, depth, number of molecules, number of labels, and diversity of datasets.
     }
\label{fig:cover_diagram_label}
\vspace{-18pt}
\end{wrapfigure}

Recent successes in language modelling \cite{gpt4, llama2} and image generation \cite{dalle3, stablediffusion} are driven by the increasing amount of training data and computational resources. Across different domains, practitioners have observed a direct relationship between model parameter count and performance on novel tasks \cite{scalinglaws}. In natural language processing, large Transformer-based models have demonstrated impressive generalization capabilities utilizing a causal autoregressive objective~\cite{gpt2}. In the meantime, image generation has undergone incredible leaps with large models trained utilizing pixel level unsupervised objectives.

While data power law scaling behavior has been tremendously beneficial in language and image domains, its practical impact on molecular reasoning and drug discovery has remained limited.
This is a direct consequence of complex scientific tasks requiring reasoning regarding the underlying structure of the data \cite{sparks}. In the past, molecular property prediction approaches have made use of graph-based methods, as these allow us to reason about the structure and interaction of different components of a molecule. Molecules are naturally represented as graphs, where the nodes represent atoms and edges correspond to covalent bonds between the atoms.

Graph Neural Networks (GNNs) have emerged as a promising way of learning molecular representations \cite{gfound, knowledge}. GNN architectures are equipped with the flexibility of learning molecular structure while building general, powerful representations over graphs utilizing backpropagation. These representations have been utilized in various paradigms such as reasoning about drug properties \cite{3dinfomax}, target binding interactions \cite{equibind}, retrosynthesis of reagents \cite{retrosynthesis}, ligand-based docking \cite{diffdock} and in-silico experiments \cite{molgnnbook}.  

Despite the growing applicability of GNNs in molecular tasks, the lack of supervised data has significantly hindered our ability to proportionally scale model sizes. It remains unclear whether graph-based architectures hold the promise of scale, similar to the paradigms of language and unsupervised image generation.

Learning molecular properties with GNNs presents its own set of unique challenges. First, multiple different GNN architectures are being actively researched. These include graph-convolutions \cite{kipf2017_gcn}, message passing architectures \cite{beaini2021directional_dgn}, graph Transformers \cite{ying2021_graphormer} and hybrid architectures \cite{rampavsek2022_gps, gps++}. These approaches have shown recent progress, but their applicability to practical applications remains an open question \cite{largemolecule}. 

Second, commonly used self-supervised training techniques do not transfer well when applied to molecular graphs; e.g., retrieving masked bonds and atoms is not informative. This is primarily due to large data requirements and the fact that graphs are limited in capturing domain-specific aspects such as chemical interactions and biological compositions \cite{gssl}. Other methods such as GPSE \cite{gpse} solely learn the graph structure, thus providing a better positional encoding for another GNN.

Lastly, public datasets have insufficient high-quality data for effective GNN architecture training. While recent attempts have been made to expand open-source datasets \cite{graphium}, their extensions towards multi-task settings remain an open question.

We aim to address these limitations and provide a concrete understanding of the required data and architectures to build foundational models for molecular graphs. Specifically, we want to answer the question: \textit{How do graph-based architectures scale in supervised multi-task training on molecular graphs?}

As summarized in Figure \ref{fig:cover_diagram_label}, we aim to answer the above question by studying the scaling behavior of 2D molecular GNNs under different settings of width, depth, number of molecules, number of labels, and the diversity in datasets. We analyze message-passing networks, graph Transformers, and hybrid architectures on the largest public collection of 2D molecular graphs. The models are tested in 2 different settings: (1) randomly split train and test sets for pretraining and (2) finetuning/probing of pretrained models on standard benchmarks.

Our work aims to provide a proper understanding of how different GNN architectures scale for molecular GNNs and its affects on performance in various settings. Our main contributions are:
\begin{itemize}[itemsep=0pt,leftmargin=*]
    \item We study the scaling behavior of 2D molecular GNNs under varied settings of depth, width, number of molecules, number of labels, the diversity in dataset, and the architectural choice. 
    
    \item We show that our largest 3 billion parameter models continue to scale with constant gains in molecular property prediction. To the best of our knowledge, this is the first work to demonstrate the continuous scaling behavior of molecular GNNs.
    
    \item We show that supervised pretraining over molecular graphs provides a rich fingerprint embedding, useful for MLP probing, and more expressive as we scale the model and datasets. 
    
    \item We provide an in-depth analysis of scaling trends across different probing and finetuning strategies. Specifically, we observe that model width and number (and quality) of labels are the most important factors driving finetuning performance.

    \item Finally, we propose \emph{MolGPS}, a foundation model derived from our findings on how to best scale molecular GNNs. MolGPS constitutes the most dominant model across the presented benchmarks to date, establishing state-of-the-art (SOTA) on 26/38 highly competitive downstream tasks.
\end{itemize}

%% file: sections/002_related_work.tex

\section{Preliminaries}

\subsection{Graph Neural Networks}

Our problem setting consists of graphs of the form $\mathcal{G} = (\mathcal{V}, \mathcal{E})$ where $\mathcal{V}$ denotes the set of nodes and $\mathcal{E}$ denotes the set of edges. Each node $i \in \mathcal{V}$ indicates the atom and each edge $(u, v) \in \mathcal{E}$ denotes a chemical bond between two atoms $u,v$. Total number of atoms in the molecule are $N = |\mathcal{V}|$ while total number of edges are $M = |\mathcal{E}|$. Node features in layer $l$ are denoted by $\mathbf{x}_{i}^{l} \in \mathbb{R}^{d_n}$ and are concatenated into an $N \times d_n$ representation matrix $\mathbf{X}^{l} = [\mathbf{x}_{1}^{l}; \mathbf{x}_{2}^{l}; . . . \mathbf{x}_{N}^{l}]^\top$. Edge features $e_{uv}^{l} \in \mathbb{R}^{d_e}$ are concatenated into the $M\times d_e$ edge feature matrix $E^{l} = [e_{uv}^{l}: (u,v) \in \mathcal{E}]^\top$.

\subsection{Scaling Laws}

We denote the parameters of a model as $\theta$ with the total number of trainable parameters being $|\theta|$. We consider a training dataset $\mathcal{D}$ consisting of labeled data samples $(\mathcal{G}, y) \in \mathcal{D}$. Here, $\mathcal{G}$ indicates the input graph and $y \in \mathbb{R}^{N}$ denotes the categorical or continuous label. Total size of the dataset is denoted as $|\mathcal{D}|$. Given the study of large foundational models, we note that $|\theta|$ is large in size and $\theta$ lies on a high dimensional manifold such that $\theta \in \mathbb{R}^{B}$ where $B >\hspace{-0.4em}> |\mathcal{D}|$. Recent work has shown that increasing the size of dataset $|\mathcal{D}|$ or the number of trainable parameters $|\theta|$ has a direct power law relationship on the loss function $L_{\theta}(|\mathcal{D}|)$ \cite{scalinglaws}. Mathematically, we have the following,
\begin{equation}
L_{\theta}(|\mathcal{D}|) \propto (|\theta_{C}|/|\theta|)^{\alpha} \label{eq:one}
\end{equation}

Equation \ref{eq:one} denotes the power-law relationship between the number of trainable parameters and the loss obtained when utilizing the parameters $\theta$. Further, $\theta_{C}$ denotes the critical parameters and $\alpha \in \mathbb{R}$ is a scalar constant. Intuitively, as the number of parameters approaches a critical value, with every gradient step, the test loss decreases at power-law with a constant rate. A similar relationship holds for the size of datasets. Mathematically, we have the following,
\begin{equation}
L_{\theta}(|\mathcal{D}|) \propto (|\mathcal{D}_{C}|/|\mathcal{D}|)^{\beta} \label{eq:two}
\end{equation}

Equation \ref{eq:two} describes the power-law relationship between the size of dataset and loss when training the model on $\mathcal{D}$. Here, $|\mathcal{D_{C}}|$ denotes the critical size of the dataset and $\beta \in \mathbb{R}$ is a scalar constant.

%% file: sections/003_methodology.tex

\section{How Do Molecular GNNs Scale?}
\label{sec:methodology}

Our study aims to answer the question: \textit{How do molecular GNNs scale?} We begin by studying GNNs in the multi-task supervised pretraining setup. Since our analysis consists of multiple tasks on a large scale, we utilize the \texttt{Graphium} library \cite{graphium}. Due to the absence of a unified consensus on the best architecture for molecular GNNs, we focus our efforts on three specific models. We select MPNN++ \cite{gps++} which improves quantum prediction over the MPNN \cite{neuralmessagepassing}, Graph Transformers \cite{attendinggt}, and Hybrid GPS++ \cite{gps++} along with the use of positional encodings. Finally, we evaluate our models on a range of public benchmarks with 38 datasets from TDC \citep{tdc}, Polaris\footnote{PolarisHub: \href{https://polarishub.io/}{https://polarishub.io/}}, and MoleculeNet~\citep{moleculenet}. We evaluate our models in both finetuning and probing (fingerprinting) settings.

We begin by providing a detailed description of the datasets and benchmarks. We then elaborate on the choice of architectures. Finally, we discuss finetuning and probing strategies along with the results of our analysis.

\subsection{Datasets}
\label{subsec:datasets}

We study the scaling behavior of GNNs on the LargeMix dataset mixture \cite{graphium}. These datasets cover different types of molecules exhibiting variable properties. Thus, the training is done in a multi-task setting consisting of thousands of labels. This is a challenging approach towards learning representations with GNNs, especially as some labels are very imbalanced and sparse. 

\textbf{LargeMix} dataset mixture consists of 5 million molecules grouped into 5 different tasks at different graph levels, with each task having multiple labels. The diversity of this mixture of data makes this dataset suitable for pretraining large GNNs. Below is a description of the individual tasks contained within the LargeMix.

\begin{itemize}[itemsep=0pt,leftmargin=*]
    \item \textbf{L1000\_VCAP and L1000\_MCF7} are two datasets of 16k and 20k molecules, respectively, with 998 graph-level classification labels corresponding to transcriptomics changes in the cell when exposed to drugs.

    \item \textbf{PCBA\_1328} is a dataset of 1.6M molecules with 1,328 binary classification labels. Each label corresponds to the activity tags of the molecules in a bioassay reported on PubChem \cite{pubchem2023}.
    
    \item \textbf{PCQM4M\_G25 and PCQM4M\_N4} are two datasets of 3.8M molecules with 25 graph-level labels and 4 node-level labels. Labels are obtained using density functional theory (DFT) simulations, a highly accurate quantum simulation method \cite{dft}.
\end{itemize}



\subsection{Finetuning and Probing Benchmarks}
\label{subsec:benchmarks}

A major benefit of foundational models is that they allow to easily generalize to unseen downstream tasks through approaches like finetuning or (linear) probing. In this work we want to also study the effect of scaling of the pretrained models on the performance on downstream tasks. For downstream task evaluation we use open-source therapeutic benchmarks. For a fair and comprehensive evaluation, all models are first pretrained using a common supervised learning strategy and then finetuned (or \emph{probed}) for molecular property prediction. The benchmarks used for evaluating are listed below.

    \textbf{TDC.} Therapeutics Data Commons~\citep{tdc} is one of the common benchmarks for drug discovery. Our study focuses on 22 ADMET (Absorption, Distribution, Metabolism, Excretion and Toxicity) tasks. While TDC serves as the bedrock for open-source drug discovery evaluation, we note that it suffers from data collection and processing biases across dissimilar molecules \cite{WeNeedBetterBenchmarks2023}.
    
    
    \textbf{Polaris.} This is a recent collection of benchmarks addressing concerns over previous datasets. Developed by an industry consortium of various biotech and pharmaceutical organizations, it provides access to high-quality molecular samples across various tasks. Our analysis considers 12 of the top tasks from either ADME (Absorption, Distribution, Metabolism, and Excretion) or DTI (Drug-Target Interaction) group for molecular property prediction.

    \textbf{MoleculeNet.} This is a benchmark dataset for molecular machine learning that is built upon public datasets \cite{moleculenet}. It consists of various datasets covering different levels of molecular properties spanning from properties at the molecular level to broader impacts on the human body. There are different categories of properties including quantum mechanics, physical chemistry, biophysics, and physiology. We investigate 4 datasets that are commonly used in similar studies such as \citep{graphmvp,graphlog,gpse}.

\subsection{Architectures}
\label{subsec:architectures}

We broadly study three types of architectures; (1) message-passing networks, (2) graph Transformers, and (3) hybrid models. In the case of message-passing networks, we focus on the MPNN++ model as it provides a suitable testbed for evaluating molecular graphs while maintaining performance across various tasks. Our graph Transformer and hybrid models make use of GPS++ model, which is known for its scalable nature on quantum property predictions. 
In addition to GNN models, we make use of Positional and Structural Encodings (PSEs) to improve the expressivity of MPNNs and introduce a soft bias into the Transformer. We discuss architectures and their design aspects below.

\textbf{MPNN++.} This is a variation of the neural message passing architecture with edge and global features \cite{neuralmessagepassing, relational, geometric}. Choosing the MPNN++ allows us to maximize architecture expressivity while minimizing the risk of overfitting on larger datasets \cite{gps++}.
Each MPNN block makes use of sequential \texttt{Dropout} \cite{dropout}, \texttt{MLP} and \texttt{LayerNorm} \cite{layernorm} modules followed by a skip connection \cite{resnet, highway} across node and edge features:
\begin{equation}
    \bar{E}^{l}, \mathbf{X}^{l} = \texttt{Dropout}(\texttt{MLP}([\mathbf{X}^{l} | E^{l}])); \quad
    \mathbf{X}^{l} = \texttt{LayerNorm}(\texttt{Dropout}(\mathbf{X}^{l})) + \mathbf{X}^{l}; \quad
    E^{l+1} = \bar{E}^{l} + E^{l} \nonumber
\end{equation}

\textbf{GPS++.} This is a hybrid model leveraging the MPNN++ inductive bias while providing the flexibility of self-attention-based modules \cite{biasedattention} to allow for a rich feature extraction scheme across nodes and edges, and was empirically proven very successful \cite{gps++}. 
Here, the standard self-attention weights are biased by a structural prior $\mathcal{B}$ from the input graph. 
Mathematically, the GPS++ module carries out the following computation:
\begin{equation}
    \mathbf{X}^{l+1}, E^{l+1} = \texttt{MPNN++}(\mathbf{X}^{l}, E^{l}); \quad
    \mathbf{Z}^{l} = \texttt{BiasedAttn}(\mathbf{X}^{l+1}, \mathcal{B}); \quad
    \mathbf{X}^{l+1} = \texttt{MLP}(\mathbf{X}^{l+1} + \mathbf{Z}^{l}) \nonumber
\end{equation}

\textbf{Transformer.} This model is identical to GPS++, but without the MPNN++ module and concatenation.
Instead, it relies solely on positional and structural encodings (PSEs) for structural bias.

\textbf{PSEs.} These are important design choices when training GNN architectures \cite{rampavsek2022_gps, gpse}, as they allow each node to understand its position and surroundings within a graph. This is essential for any graph Transformer, but it was also shown to improve the expressivity of molecular GNNs.
Specifically, we use three PSE schemes. First, we use random-walk diagonals \cite{dwivedi_lpse} to allow one to decouple structural and positional representations. Learned positional encodings are used to tackle isomorphic nodes. 
Second, we use Laplacian eigenvectors \cite{beaini2021directional_dgn} as these form an expressive way to encode node geometries and positions. Laplacian eigenvectors provide strong theoretical guarantees with respect to the expressivity of the Weisfeiler-Lehman test, a useful insight when evaluating GNNs at scale. 
Last, we use the Laplacian eigenvalues \cite{san} as a suitable PSE scheme to fully leverage the Laplacian spectrum. 
Additionally, they provide global structural information about the graph.

\subsection{Finetuning and Probing}
\label{subsec:fine_tuning}


Following pretraining, we finetune our pretrained models on a range of unseen downstream tasks. While there exist no clear guidelines for finetuning GNNs, this aspect is extensively explored in this work. Notably, our evaluation considers two strategies (finetuning and probing), which both significantly benefit from increased scale of the pretrained model.

\textbf{Finetuning.} Since our training setup consists of multiple tasks and our architectures incorporate multiple task heads, we need to identify a \textit{finetuning module} after which the remaining pretraining architecture is removed and replaced by a newly initialized \texttt{MLP}: the \textit{finetuning head}. As all downstream tasks discussed above reside on the graph level, our main choice is the \textit{graph output network}, the \texttt{MLP} that processes features after being aggregated from the node to the graph level, and further feeds into the task heads for graph-level tasks. Intuitively, this layer's output representations have benefited the most from pretraining on diverse data and tasks, as it feeds directly into the various task heads. We further investigate the effect of choosing layers of the tasks heads as finetuning module to potentially leverage synergies between specific pretraining and downstream tasks. As all downstream tasks are on the graph level, we trim the architecture by removing parts related to node level tasks and unused task heads.




\textbf{Probing and Fingerprinting.} Similar to finetuning, we employ probing using an \texttt{MLP} as a strategy for solving new downstream tasks. For probing, the base model is kept frozen and only the new layers are trained. This allows the training procedure to focus the gradient on newly added parameters, resulting in task-specific probing head layers. In the case of large model sizes, extracting embeddings (so-called \emph{fingerprints}) from the frozen base model is expensive with respect to memory consumption and compute. We tackle this bottleneck by caching fingerprints on disk and reusing them during probing. Since the gradient does not impact parameters of the base model, fingerprints remain unchanged yielding an optimized strategy for downstream tasks capable of parallelization across multiple inexpensive devices. In this work, similar to the finetuning setup, we extract fingerprints from the task head \texttt{MLP}s of graph level tasks, and from the \textit{graph output network}, the \texttt{MLP} that directly feeds in to the task heads.


%% file: sections/004_scaling_laws.tex

\section{Experiments}
\label{sec:scaling_laws}

\subsection{Scaling Trends for  Pretraining}
\label{subsec:experiments-scaling-pretrain}

\begin{wrapfigure}{R}{0.55\columnwidth}
\centering
    \includegraphics[width=\linewidth]{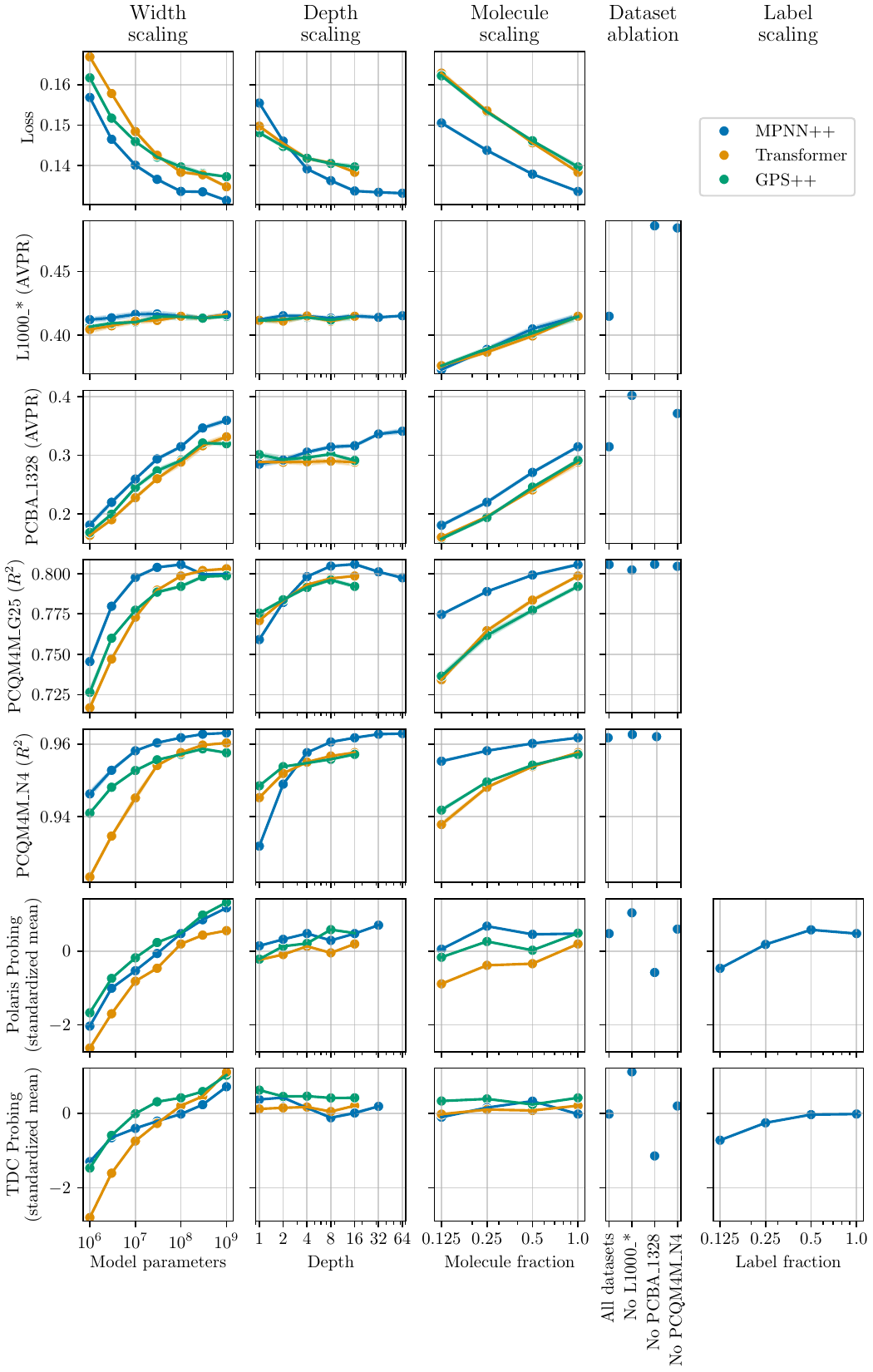}
    \caption{Effect of scaling different scaling types (columns) to test performance (rows). The \textit{standardized mean} is calculated as mean of standardized scores for every task in a dataset group, i.e., a mean and standard deviation per task were calculated based on all our models in this study (signs of tasks with \textit{lower is better} metrics were flipped).}
\label{fig:pretraining_one}
\end{wrapfigure}

In this section, we evaluate the scaling behaviour of our models according to various parameters summarized in Figure \ref{fig:cover_diagram_label}, namely the architectural choice, width, depth, number of molecules, labels, and different datasets. 
We analyze our models on datasets from LargeMix described in Section~\ref{subsec:datasets}. For detailed results and experiments of our study, please refer to the supplementary material.

\textbf{Overview.} Figure \ref{fig:pretraining_one} presents the variation of architectural choices between MPNN++, Transformer and GPS++, with training curves in Appendix \ref{sec:additional_pretraining}, and full finetuning (and probing) results in Appendix~\ref{sec:additional_downstream_results}. Notably, all models scale favourably with the increasing scale of width (number of parameters per neuron), depth (number of GNN layers) and number of molecules (dataset size). 

\textbf{MPNN++ vs Transformer.} MPNN++ models are more parameter efficient as they perform better with small width and depth compared to Transformers. They are also data efficient as they perform significantly better for the quantum PCQM4M\_* tasks when sub-sampling the datasets, although smaller differences are observed for the remaining (biological) datasets. Transformers being ``data-hungry'' is consistent with recent works in domains such as natural language and computer vision \cite{gpt, flamingo, knowledge}. 
The hybrid GPS++ seems to benefit from the MPNN++ expressivity in low-parameter regimes, while also exhibiting a similar molecular scaling to the Transformer in low-data regimes.
Finally, we notice that MPNN++ models are more affected by depth, an unsurprising outcome considering that, contrarily to Transformers, their receptive field depend on the number of layers.

\textbf{Width scaling.} As seen in the first column of Figure \ref{fig:pretraining_one}, increasing the width has a significant impact on model performance across all tasks. Further, we trained larger models for fewer epochs as the loss converged faster and more likely to exhibit overfitting on the PCQM4M\_* tasks.

\textbf{Depth scaling.} Similar to width, depth of GNN models plays an important role in the dataset fit in test time. Deeper models with larger layers capture intricate aspects of the data resulting in 12.5\% improvement in test error. However, performance plateaus at around 8-16 layers for quantum datasets. For PCBA\_1328, the performance continues to increase. 

\textbf{Molecule scaling.} Unsurprisingly, the number of molecules in the training set correlates strongly with the performance of all models. Contrary to width and depth, molecule scaling is consistent across all models and test sets, with GPS++ models and Transformer benefiting more than MPNN++ on quantum tasks. 
For instance, increasing the dataset size by eight-fold (12.5\% to 100\%) yields a significant 33.33\% improvement in model performance in the case of the hybrid GPS++ model.

\textbf{Detailed scaling law analysis.} We provide detailed analysis of the observed scaling trends in terms of Equations~\ref{eq:one}~and~\ref{eq:two} in Appendix~\ref{sec:app:scaling-laws}, observing scaling laws similar to those in other domains~\cite{scalinglaws}.

\subsection{Scaling Trends on Downstream Tasks}
\label{subsec:experiments-scaling-downstream}


We now evaluate scaling of models when finetuning and probing on downstream tasks. As detailed in Section \ref{subsec:fine_tuning}, all weights are tuned in the case of finetuning, while the pretrained model is frozen when fingerprinting followed by probing. 
\begin{wrapfigure}{R}{0.5\columnwidth}
\centering
    \includegraphics[width=\linewidth]{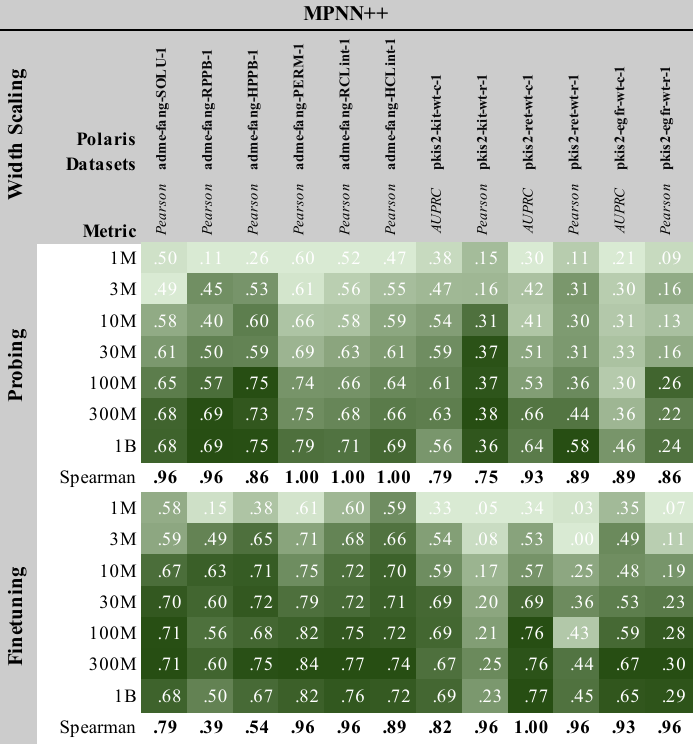}
    \caption{Finetuning and probing performance of pretrained MPNN++ models of different width on the \underline{Polaris} benchmark. \textbf{\color{dark2green}{Darker green}} shades denote better metric values. Larger models tend to perform better on unseen tasks. Spearman correlation values closer to 1 indicate that predictive performance correlates with larger model sizes.}
\label{fig:polaris_width_MPNN}
\end{wrapfigure}

Due to the large number of downstream tasks spread across 38 tasks, we limit our evaluation to probing for most experiments, except for MPNN++ where we also finetune the model.

To summarize scaling trends, we compute the Spearman's rank correlation coefficient \cite{spearman} between model performance on a given metric and the scale of model/data used. The correlation is given by a value in the range $ [-1, 1]$ , with a value of \textit{$1$} indicating perfect scaling (i.e., a larger model or dataset yields better downstream task performance), \textit{$-1$} indicating imperfect scaling (i.e., a smaller model or dataset would be preferred) and \textit{$0$} indicating no correlation. We note that this evaluation scheme, although statistical, aims to answer an important question: \textit{What kind of design decisions are necessary to build a foundational model for molecular representations?}

\textbf{MPNN++ vs. Transformer.}
For probing on downstream tasks, we study the effect of architecture choices of width, depth, and number of molecules.
We find that Transformers benefit more from increased width on downstream tasks compared to GPS++ and MPNN++ as seen in Figure~\ref{fig:pretraining_one} (bottom two columns).
Despite the number of molecules having a stronger impact on all model's performance, it only slightly impacts the downstream performance of all models, with a small benefit for MPNN++.
Finally, Transformer is the only model with a \textit{small} positive trend for depth scaling, while GPS++ and MPNN++ show close to no trend (Figure~\ref{fig:pretraining_one}).

\begin{figure*}[t]
    \centering
    \makebox[\textwidth]{%
        \includegraphics[width=1.05\textwidth]{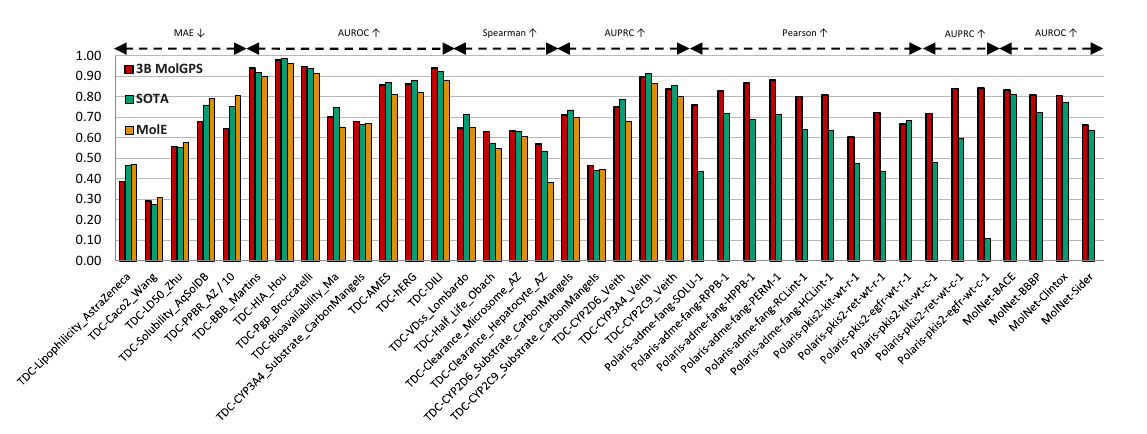}
    }
    \caption{Comparison of our MolGPS foundation model (that combines fingerprints from the MPNN++, Transformer and hybrid GPS++ model) to the SOTA across \underline{TDC}, \underline{Polaris}, and \underline{MoleculeNet} benchmarks. SOTA refers to the maximum value for each dataset.
    MolGPS establishes new SOTA on \textit{$11/22$} TDC tasks and on all but one task among Polaris and MoleculeNet.
    }
    \label{fig:sota_comparison}
    \vspace{-10pt}
\end{figure*}

\textbf{Width scaling.} We evaluate width scaling on Polaris and TDC datasets in Figure 
\ref{fig:polaris_width_MPNN} (and Figure~\ref{fig:TDC_width} provided  in the appendix). We observe linear scaling trends for MPNN++ on all Polaris datasets, with an average Spearman correlation of 0.91 for probing and 0.85 for finetuning. On TDC, a similar trends are observed (on average 0.69 for probing and 0.72 for finetuning) with a strong correlation of \textit{$>\!0.75$} for \textit{$15/22$} datasets during probing and \textit{$17/22$} during finetuning. These results strongly indicate the benefits of larger pretrained GNNs for downstream tasks, a result consistent with prior findings in scaling of large models \cite{scalinglaws}. Similarly, Transformer and GPS++ show strong positive scaling trends with model width.

\textbf{Depth scaling.}
We evaluate the scaling of depth of MPNN++, GPS++ and Transformer models on the Polaris and TDC benchmarks in Figures \ref{fig:polaris_depth_MPNN} and~\ref{fig:TDC_depth_MPNN} (Appendix~\ref{subsec:app:depth_scaling}). For probing on Polaris, we observe weak positive trends, with average scaling Spearman correlations of \textit{$0.47$}, \textit{$0.55$}, and \textit{$0.50$}, respectively. We see weaker average correlations on TDC, being slightly negative for MPNNN++ and GPS++ and best for Transformer with \textit{$0.27$}. However, finetuning MPNN++ achieves a respectable correlation of \textit{$0.33$}. 
While some datasets strongly benefit from deeper networks, others strongly deteriorate with no clear pattern observable for the TDC datasets.
We conjecture that degradation with depth is related to the oversmoothing issue described in Appendix~\ref{sec:oversmoothing}. 
Certain molecular properties can be well predicted only from small local substructures, hence eliminating the need for long-range interactions that deeper networks enable.

\textbf{Molecule scaling.} 
In this setting, we randomly sub-sample a number of molecules in the training set by 12.5\%, 25\% and 50\% to study their effect on downstream tasks.
Surprisingly, probing and finetuning performance does not correlate strongly with the amount of molecules in the training set, as reported in Figures \ref{fig:polaris_molecule} and~\ref{fig:TDC_molecule_MPNN} (Appendix~\ref{subsec:app:molecule_scaling}). For MPNN++, we observe average Spearman correlations of \textit{$0.28$} and \textit{$0.32$} when probing and finetuning on TDC, respectively.
Contrarily to their stronger trends on the pretraining tasks, Transformer and GPS++ have lower correlations during probing of \textit{$0.13$} and \textit{$0.15$}. In the case of Polaris, only average correlation of Transformers stands out at \textit{$0.73$}, however reaching worse peak performance per task compared to the less correlated MPNN++ and GPS++. The globally weak positive trends come from the variation across the downstream tasks, with many strong correlations and a few strong negative correlations.

\textbf{Label scaling.}
We now study the effect of target labels by randomly sub-sampling the number of labels of each dataset in the training set by 12.5\%, 25\% and 50\%.
In Figures \ref{fig:polaris_label_MPNN} and \ref{fig:TDC_label_MPNN} (Appendix~\ref{subsec:app:label_scaling}), we observe a large Spearman correlation of \textit{$0.57$} on Polaris and \textit{$0.54$} on TDC between the ratio of training labels and the performance, with only a few negative correlations. In the finetuning regime, this number lowers to \textit{$0.37$} on TDC.
These stronger correlations put \textit{label scaling} as the second-best strategy for improving the model's downstream performance.

\textbf{Dataset ablation.}
We further conducted a study to determine the importance of the pretraining data in two ways. Firstly, we repeat pretraining of the models without specific pretraining datasets (\textit{dataset ablation}).
Secondly, we probe models specifically from certain task head MLPs compared to the base GNN models (\textit{task-head ablations}).

Observing the dataset ablations in Figure \ref{fig:Polaris_dataset_MPNN} and~\ref{fig:TDC_dataset_MPNN} (Appendix~\ref{subsec:app:dataset_ablation}), we see that PCBA\_1328 is the most important pretraining dataset for downstream task performance while L1000\_* actually hurts the performance on certain tasks. It will therefore prove beneficial to pretrain without the L1000\_* datasets as we will see later.

\textbf{Task-head ablations.} We further tested the effect of probing from different layers of the task heads rather than the graph output network. Results are shown in Figure~\ref{fig:TDC_POLARIS_task_head_mpnn} (Appendix~\ref{subsec:app:task_head_ablation}). While overall, the graph output network leads to best performance and correlation, the representation after the first layer of the PCBA\_1328 task head performs strikingly well for some tasks, possibly due to synergies from pretraining on bioassays. This suggests probing approaches using combinations of fingerprints could further improve results. On the other hand, the layers from the PCQM4M\_G25 dataset perform poorly, which is intuitive as this pretraining task is dissimilar to the downstream task.

\textbf{Probing vs. finetuning.}
So far, we have considered finetuning and probing side-by-side, establishing both as effective strategies for tackling downstream tasks. However, given the relatively similar performance and the significantly higher computational cost of finetuning, we find probing to be the overall more advantageous approach. Another major benefit of probing is the ability to leverage multi-level information from the pretrained GNN as investigated in our task head ablation study above. We recall that our pretraining is based on a supervised multi-task learning approach. As a result, different task heads capture task-specific information, while earlier layers that feed into the task heads carry more general information. When we combine fingerprints from various layers, we can think of taking into account knowledge from several “experts”.



\subsection{Towards a Final Foundation Model}
\label{subsec:towards_a_final_foundation_model}

We now explain how the above findings can be pieced together to develop \emph{MolGPS}, a powerful graph foundation model. Apart from scaling the model width, we found two other design choices with a major impact on the performance for the various downstream tasks. We report results on the MoleculeNet benchmark~\citep{moleculenet} here in addition to the previously used TDC and Polaris benchmarks.


\textbf{Multi-fingerprint probing.}
Our previous task-head ablation study suggested that different fingerprints may be optimal for probing depending on the downstream task. As a result, we choose probing (instead of finetuning) and further experiment with combinations of multiple fingerprints extracted at different layers of a pretrained model, which improves performance on downstream tasks. Moreover, performance can be further enhanced by combining fingerprints from multiple pretrained models.

\textbf{Pretraining without L1000.}
Additionally, based on our observation in the dataset ablation, we pretrained new models without the L1000\_* pretraining tasks, which leads to performance improvements across all scales. We hypothesize this is due to the challenging signal-to-noise ratio for those particular tasks, as also pointed out in the literature~\citep{tong2023transigen}.

\begin{wrapfigure}{}{0.65\columnwidth}
\vspace{-14pt}
\centering
    \includegraphics[width=\linewidth]{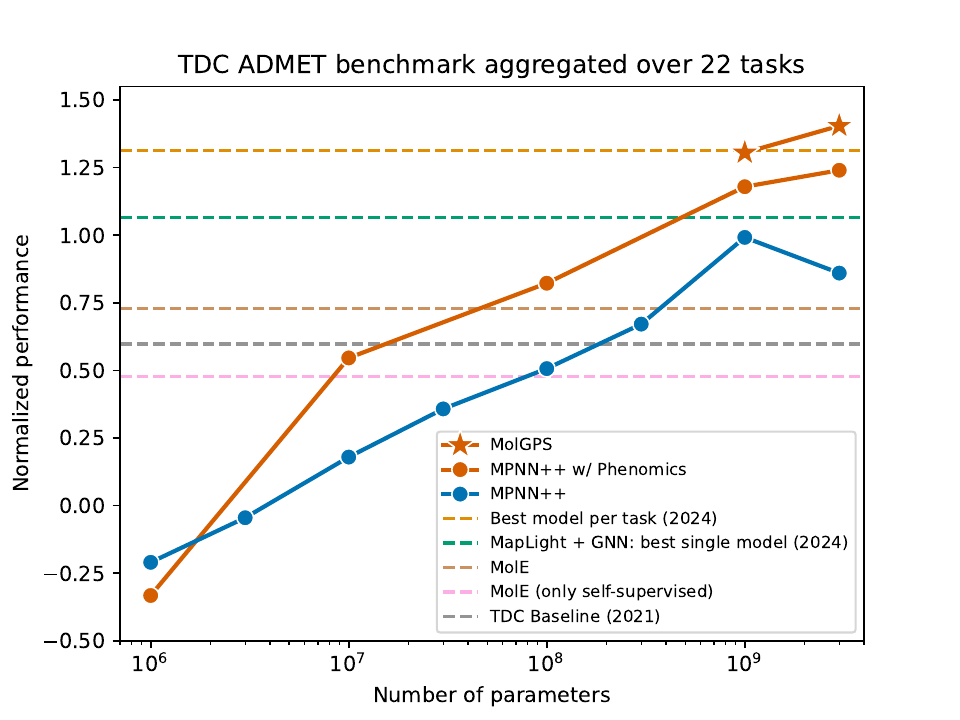}
    \caption{Comparison of our MPNN++ probing (that leverages multiple fingerprints; with and without additional phenomics pretraining) and MolGPS (that leverages fingerprints from MPNN++, Transformer and GPS++) to various baselines across \underline{TDC} benchmark collection using an aggregated metric.}
    \label{fig:sota_aggregated}
\end{wrapfigure}
In Figure~\ref{fig:sota_aggregated}, we present multi-fingerprint probing results of the MPNN++ model. We report an aggregated metric across the TDC benchmark collection (normalized performance\footnote{
The normalized performance of a model on a benchmark collection is calculated by first deriving the z-scores of the model's performance per task relative to the leaderboard. We flip the sign of z-scores for tasks that are ``the lower the better`` and then average the z-score across the tasks.}).
We observe a strictly positive scaling trend up to 1B parameters, clearly outperforming TDC Baseline (the normalized score across the best models per task reported in~\citep{tdc}) and the MolE foundational model~\citep{mole}, a gold standard for molecular property prediction (including a model variant with \emph{only} self-supervised pretraining). The 1B MPNN++ only slightly trails MapLight+GNN~\citep{notwell2023admet}, the best-performing \emph{single} model on the TDC benchmark. Surprisingly, we were unable to further scale the model to the 3B parameter regime, likely due to limitations of our pretraining data mix.

\textbf{Integrating phenomics into pretraining.}
To improve our pretraining data mix, we experimented with an additional data type for model pretraining that comes from phenomic imaging. The dataset contains ~6k labels for ~500k molecules that were derived from phenomic imaging~\citep{bray2016cell} of cells perturbed with either a dose of a compound or a gene knockout. We conducted a similarity analysis between the obtained images (represented by vector embeddings, e.g., similar to~\citet{he2022masked}) subject to a compound perturbation on one side, and images subject to a gene perturbation on the other side. The pretraining task is to predict for each compound if it has a phonemically visible similarity to a gene knockout (indicating a biological relationship).

Adding phenomics data to our pretraining data mix (i.e., PCBA\_1328 and PCQM4M), improved our downstream task performance across the board. Comparing the scaling trends in Figure~\ref{fig:sota_aggregated}, MPNN++ with phenomics pretraining exhibits a significant vertical upwards shift compared to the original MPNN++. Notably, we were also able to extend our scaling study to the 3B parameter regime. While we were previously unable to extend the scaling trend, MPNN++ with phenomics maintains a positive scaling trend.

\textbf{MolGPS.}
We introduce a final graph foundation model that leverages the various findings of this paper. MolGPS inherits many architecture design choices from the \textbf{G}eneral, \textbf{P}owerful, \textbf{S}calable Graph Transformer method~\citep{rampavsek2022_gps} and can be used to \emph{navigate} the molecular space.
Our 1B and 3B MolGPS combines fingerprints from MPNN++, Transformer, and GPS++ models (of scale 1B and 3B, respectively) that have been pretrained \emph{with} the new phenomics data and \emph{without} the L1000 dataset, followed by a specialized probing \texttt{MLP}. Figure~\ref{fig:sota_comparison} compares this model across the TDC, Polaris and MoleculeNet benchmark collections to the current SOTA for each task and to MolE. MolGPS yields by far the strongest downstream task results, outperforming MolE in \textit{$21/22$} TDC tasks and establishing SOTA performance on \textit{$11/22$} TDC tasks. This makes MolGPS the model with the most SOTA entries in the TDC leaderboard followed by MapLight+GNN~\citep{notwell2023admet} that established SOTA on 5 TDC tasks and 7 other methods that are SOTA for at least one TDC task.\footnote{We report scores from the TDC leaderboards extracted on March 15, 2024. SOTA on TDC is established by a group of 8 different models, namely Chemprop-RDKit~\citep{chemprop}, MapLight, MapLight+GNN~\citep{notwell2023admet}, BaseBoosting~\citep{baseboosting}, CFA~\citep{cfa}, SimGCN, ZairaChem~\citep{zairachem} and ContextPred~\citep{contextpred}. For MoleculeNet, we use the same split as GraphMVP~\citep{graphmvp}, thus report the best results from their table.}
Similarly, compared to previous best methods on the Polaris and MoleculeNet benchmarks, we observe that our model is significantly better (often by large margins) for all but one  downstream task. We primarily attribute the large-scale success to width scaling up to 3B parameters and the integration of phenomics data for pretraining.
Figure~\ref{fig:sota_aggregated} shows the normalized performance of MolGPS for the TDC benchmark, where it performs comparable to the best model per task for 1B parameters and clearly outperforms that baseline for 3B parameters. This is remarkable recalling again that this score is derived from the best scoring method \emph{per task} of the benchmark collection, while we use a single method for all tasks. We finally note that further comparison of MolGPS to several foundation models that rely on self-supervised pretraining can be found in Appendix~\ref{subsec:app:unsupervised}.


%% file: sections/006_discussion.tex

\section{Conclusion}
\label{subsec:conclusion}

In this paper, we studied the scalability of GNN models including message-passing networks, graph Transformers and hybrid architectures on the largest public collection of 2D molecules for the tasks of molecular property prediction.
We showed major performance gains from the growing amount of parameters, data and compute, both on the original test set and on downstream finetuning. 
Importantly, our models benefit tremendously from the increasing scale of width, number of molecules, and number of labels. 
Our largest 3B parameter models, including MPNN++, Transformer, and GPS++, continue to scale favourably.
More importantly, we demonstrate a consistent performance improvement on downstream property prediction tasks via finetuning and probing as we scale model and data size.
Finally, we derive MolGPS, a powerful foundational model based on a multi-fingerprint probing approach that can be used to navigate the chemical space, establishing state-of-the-art on 26 out of 38 highly competitive downstream tasks.
We hope that our work paves the way for the development of foundational GNNs and new architectures with applications in pharmaceutical advancements and drug discovery.
\paragraph{Future Work.} \label{par:future_work}
While our study demonstrates the benefits of increasing number of parameters far greater than prior work, there are still orders of magnitude before we reach a general-purpose foundational model of molecules. Our analysis is restricted to the effect of number of parameters and molecules during pretraining and finetuning stages. Future work would aim to uncover additional aspects of GNN training such as the increasing complexity of aggregation functions and their effect on scaling properties. It will also be important to bridge current limitations for training large GNNs for molecules related to the expensive featurization as graphs, fast data loading and methods that unlock even larger models such as model parallelism.





\paragraph{Broader Impact.}
\label{sec:broader_impact}
We foresee positive impacts of GNNs in areas of drug discovery, pharmaceutical advancements and tackling rare diseases by studying their molecular configurations.
On the other hand, such models could also be used for harmful purposes such as developing chemical weapons and biohazards. We note that the usage of GNN models for such applications is less likely.

%% file: sections/009_Appendix.tex
\appendix
\onecolumn

\section{Related Work}
\label{sec:related_work}

\textbf{Foundation Models for Molecules.} Recent work has relied on foundation models as a generalist class of models for sequential modelling \cite{power, gfound} as well as knowledge retrieval \cite{knowledge}. Within molecular drug discovery, recent works rely on structured models of ligands \cite{chemlm}. \citet{biogpt} and \citet{denovo} study a general model for protein synthesis. \citet{contactprediction} construct a self-attention driven architecture for contact prediction. \citet{functionalproteins} learn to generate a family of functional proteins. \citet{progen2} present a class of protein-pretrained language models.
Similarly, \citet{mole} study binding interactions between different assays at the molecule-molecule interaction level.
While many models focus on the design of molecules, a recent class of methods has also focused on properties of molecules \cite{graphium}. 
Our study follows up by exploring similar\textit{ molecular tasks} for property prediction.

We elaborate more on additional advancements in foundational models making use of molecular graphs. Recent works have argued that the use of high-capacity models will be a significant boon to scientific discovery tasks \cite{scientific}. Of specific interest are tasks in the quantum and molecular discovery paradigms \cite{aiquantum} which demand domain-specific expertise such as knowledge of structure, provision of additional inductive biases and large data requirements. Towards this hypothesis, \citet{incontext} present an in-context learning framework for molecular property prediction without explicitly using a meta learning procedure. This leads to a general algorithm capable of discovering high-level structures from a pretraining sample set. \citet{fewshot} propose a similar framework making use of few-shot learning techniques resulting in a sample-efficient learning procedure. \citet{graphmodelling} present an alternative approach by modelling the full graphical structure of molecules across different property prediction tasks. Although effective, modelling the entire graph results in a computationally intensive learning procedure. Finally, \citet{graphsegment} scale up learning to larger graph sizes by segmenting graph neighborhoods on the fly. An ad-hoc partitioning procedure is employed and interleaved with the learning phase in order to accelerate learning on larger and dense graphical clusters.

\textbf{Architecture Design.} Recent methods in graph architecture design focus on attending to structural information across nodes \cite{attendinggt}. Of specific interest are graph Transformer networks which extract node as well as edge information by composing sequential attention modules over graph readouts \cite{gtns}. In parallel, graph attention networks model attention weights across edges of a graph \cite{graphattention}.

While attention mechanisms have demonstrated modern progress, traditional architectures such as neural message passing \cite{neuralmessagepassing}. 
On one hand, message passing provides a rich and expressive framework for constructing representations. 
\citet{regularization} study regularization based on noisy nodes for the task of molecular property prediction. Provision of noise imputation in node-level features leads to simple and expressive method for tackling sparse molecular graphs. 
Graph bootstrapping \cite{graphbootstrapping} allows prior architectures to scale up to larger and complex graphs for representation learning. 
Our exploration of \textit{different architectures} is aligned with the aforesaid works in literature, and with recent trends towards Transformers in related machine learning fields.

\textbf{Scaling Laws.} Recent work in model scaling has demonstrated that performant models follow a power law relationship between their parameter sizes and performance on new data samples \cite{scalinglaws}. Additionally, this relationship holds during the finetuning stage \cite{finetuningscaling}, thus indicating a strong reliance on model parameters. 
\citet{explaining} explain this power law fit by observing learning as moving on a smooth data manifold. 
\citet{mixedmodal} study the power law fit for mixed modality generative models, indicating that the scaling behavior is modality agnostic across various datasets. The result hints at the generality of scaling across different domains and applications, which can be extended to the study of scaling laws towards different training regimes (such as finetuning, downstream transfer and inference) and different problem domains (vision, language, audio, generative modelling, etc.)
Our exploration of \textit{scaling behaviors in graph networks} is motivated by the aforesaid directions.

\textbf{Expressivity of GNNs.}
Prior work highlights that GNN architectures are limited in their expressivity to distinguish between graphs of similar node arrangements but different geometrical structures \cite{gin}. Various works indicate this as a consequence of aggregation functions and other design factors involved in GNN training \cite{grl}. On the other hand, recent work argues that only specific architectures are found robust to over-smoothing when building latent representations \cite{transformeroversmoothing}. For instance, graph Transformers exhibit over-smoothing robustness as they utilize strong inductive biases such as attention. \citet{extrapolate} connect the limited expressivity of GNNs with their ability to extrapolate on simpler tasks. Contrary to multi-layer networks, GNNs struggle to extrapolate on simpler tasks but show promise for improvement. \citet{weisfeiler} aim to tackle over-smoothing by building higher-order GNN architectures capable of capturing intricate node characteristics in their deeper layers. Finally, \citet{hierarchical} present the differentiable pooling module capable of pooling neighboring node features which aid in reducing noise across layer representations.

\section{Experimental Details}
\label{sec:implementation}

\subsection{Pretraining}

All models use 2-layer MLPs to encode node and edge features, respectively, followed by the core model of 16 layers of the MPNN++, Transformer or GPS++ (except for when scaling depth). To be able to tackle graph-level tasks, the outputs are aggregated to the graph level, e.g., by summing them up across the atoms of a molecule. Then, node and graph level representations go through separate 2-layer MLPs. Finally, representations are processed by separate task heads (2-layer MLPs) specific to each pretraining task. Further, all layers use layer norm and dropout with $p=0.1$. The encoder and model core additional have residual connections similar to the design in~\citet{he2016deep}.

Our hyperparameter search for all base models was conducted on all oberserved data samples with a constant model size of $10M \pm 0.1M$ parameters. For scaling on width, zero-shot scaling from $\mu P$ \cite{yang2022tensor_mup} was used. For other scaling results, $\mu P$ was used to scale the model with $10M$ parameters used as the base model. In the case of depth scaling, we adjusted the learning rate as suggested by depth-$\mu P$ \cite{yang2023tensor_mup_depth}. We did not consider adjusting the residual connections.

Our base MPNN++, Transformer and hybrid GPS++ models are trained using \textit{Adam} with a base learning rate of \textit{$0.003$}, \textit{$0.001$,} and \textit{$0.001$}, respectively. We use 5 warm-up epochs followed by linear learning rate decay. All pretraining has been conducted with a batch size of 1024. Scaled version of the used models require advanced training strategies due to the large model size. We used multi-gpu training (with up to 8 NVIDIA A100-SXM4-40GB GPUs) and gradient accumulation, while adjusting batch size to keep the effective batch size constant. Most models were trained on sigle gpus but our 300M and 1B parameter models used 4 and 8 gpus, respectively.

\subsection{Finetuning and Probing}

\textbf{Finetuning.} As outlined in Section~\ref{subsec:fine_tuning}, a finetuning module is selected from one of the layers of the pretraining architecture and a newly initialized \texttt{MLP} is appended to that layer. Here, we use 2-layer \texttt{MLP}s with a hidden dimension of 256. For each experiment, when retraining this model, we set the dropout rate to zero and train for 40 epochs using a batch size of 256 and a constant learning rate of $0.0001$. To first adjust the \textit{finetuning head} -- the newly initialized \texttt{MLP} after the finetuning module -- we freeze the remaining architecture for the first 10 epochs. To find a unified finetuning strategy for each pretrained model/downstream task combination, we select the best epoch where validation performance was maximized across all seeded runs of the experiment.

\textbf{Probing.} Similar to finetuning, we apply a 2-layer \texttt{MLP} to the fingerprints derived from the pretrained model. We choose a hidden dimension of 128 and train for 30 epochs with a batch size of 128 and a constant learning rate of $0.0001$. Further, we use the same approach as for finetuning to select a unified number of epochs for each pretrained model/downstream task combination based on validation.

We finally note that all downstream task experiments were conducted on single cpus.


\subsection{Performance Metrics}
We provide detailed explanations for the metrics used for evaluating the performance of different tasks throughout this work.
\begin{itemize}[itemsep=0pt,leftmargin=*]
    \itemsep0em
    \item \textbf{Pearson.} The Pearson correlation coefficient measures the strength and direction of the linear relationship between two datasets. It ranges from \textit{$-1$} to \textit{$+1$}, where \textit{$0$} indicates no correlation, \textit{$-1$} indicates a perfect negative linear relationship, and \textit{$+1$} indicates a perfect positive linear relationship.
    \item \textbf{Spearman.} The Spearman correlation coefficient is a nonparametric measure of the monotonic relationship between two datasets. 
    Similar to other correlation coefficients, Spearman's correlation varies between \textit{$-1$} and \textit{$+1$}, with \textit{$0$} indicating no monotonic relationship. Correlations of \textit{$-1$} or \textit{$+1$} imply a perfect monotonic relationship between the two datasets.
    \item \textbf{AUROC.} The ROC curve demonstrates the trade-off between the true-positive rate and the false-positive rate at different thresholds. The area under the ROC curve (AUROC) expresses how good a model is regardless of the chosen threshold.
    \item \textbf{AUPRC.} This metric is useful particularly when dealing with imbalanced datasets and it summarizes the area under the precision-recall curve at various thresholds.
    \item \textbf{MAE.} Absolute error is the magnitude of the difference between a prediction and the true value of an observation. Mean Absolute Error (MAE) calculates the average of these absolute errors for a group of predictions and observations, providing a measure of the overall error magnitude for the group.
    \item \textbf{Average Standardized Score.} The standardized score of a model $M$ for a specific task $T$ is derived as follows. Let $s_T(m)$ denote the score of a model $m$ for task $T$ and let $\mathcal{S}_T \coloneqq \{s_T(m): m\in\mathcal{M}_T\}$, where $\mathcal{M}_T$ denotes the set of all methods that have been applied to task $T$. The aggregated score is defined as $\text{s\_agg}_T(M) \coloneqq \text{sgn}(s_T) \cdot \left(s_T(M) - \text{mean}(\mathcal{S}_T)\right)/\text{std}\left(\mathcal{S}_T\right)$, where $\text{sgn}(s_T)$ is the polarity of the metric, i.e., positive if ``higher is better'' and negative if ``lower is better''.
\end{itemize}


\section{Trade-Off Between Over-smoothing and Depth}
\label{sec:oversmoothing}

We note that GNN architectures exhibit \textit{over-smoothing} phenomenon, which implies that latent representations of a network become similar and coarser as the network grows in size. Prior evidence suggests that over-smoothing occurs linearly with the increasing depth of GNN networks \cite{grl, gin}. 
We observed similar behaviors for MPNN architectures during pretraining where the performance for node-level tasks degrades significantly with very deep networks. However, it is difficult to determine without any doubt that over-smoothing is the culprit.

On another hand, over-smoothing is believed to be aleviated by graph Transformers.
Recent works argue that Transformers present favorable properties which make them robust towards over-smoothing, such as the provision of embeddings and the inductive bias of attention \cite{transformeroversmoothing}.  However, we still observe a degradation of performance with depth of our Transformer models, in contradiction with this hypothesis. Its theoretical understanding and empirical analysis remains an open question for future work.

\newpage
\section{Training Curves of Pretraining Models}
\label{sec:additional_pretraining}

\begin{figure}[H]
    \includegraphics[width=\textwidth]{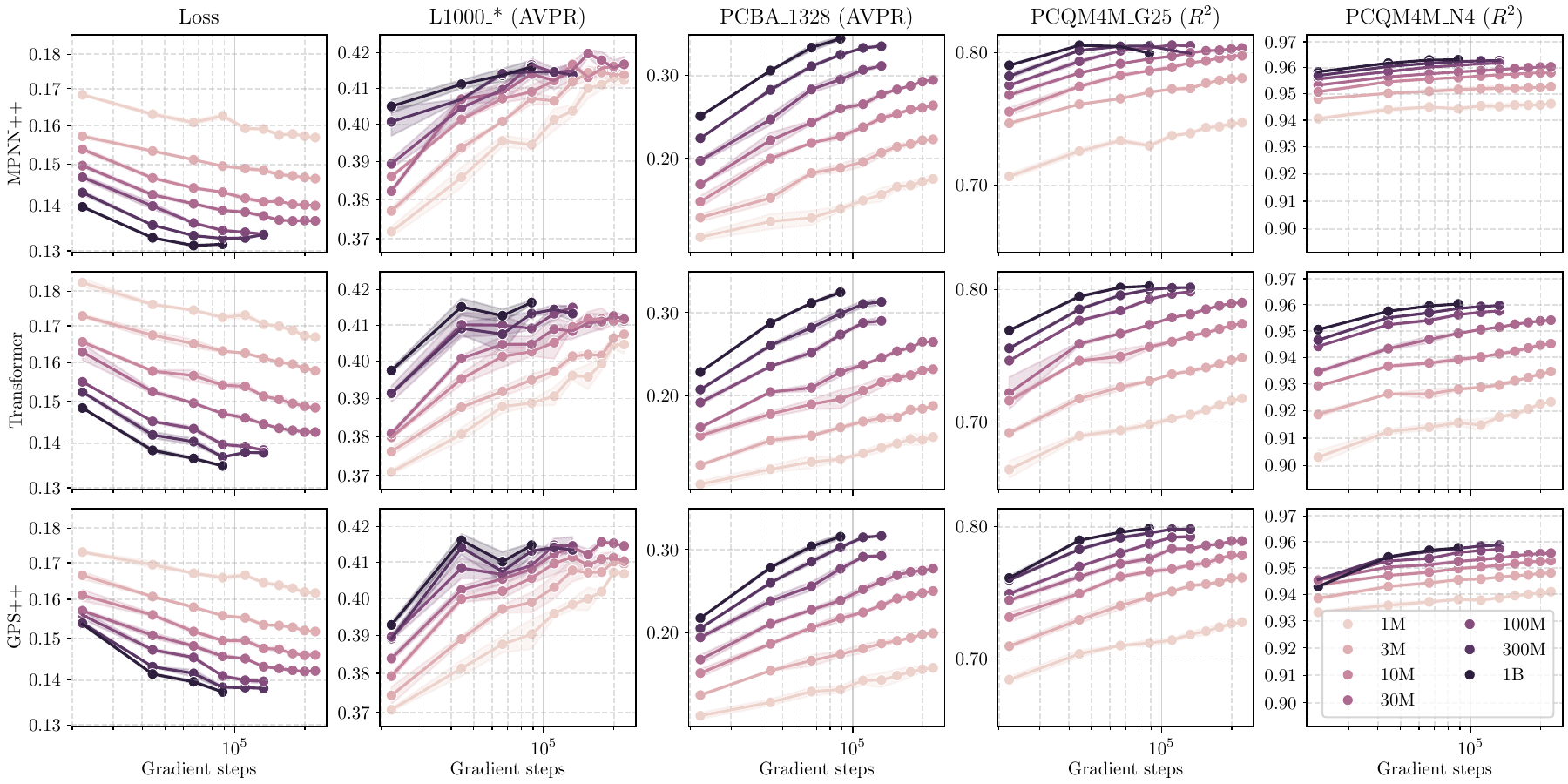}
    \caption{Model performance on the test set throughout training for MPNN++, Transformer, and GPS++ architectures with width scaling. Different colors represent models with varying number of parameters.}
\end{figure}

\begin{figure}[H]
    \includegraphics[width=\textwidth]{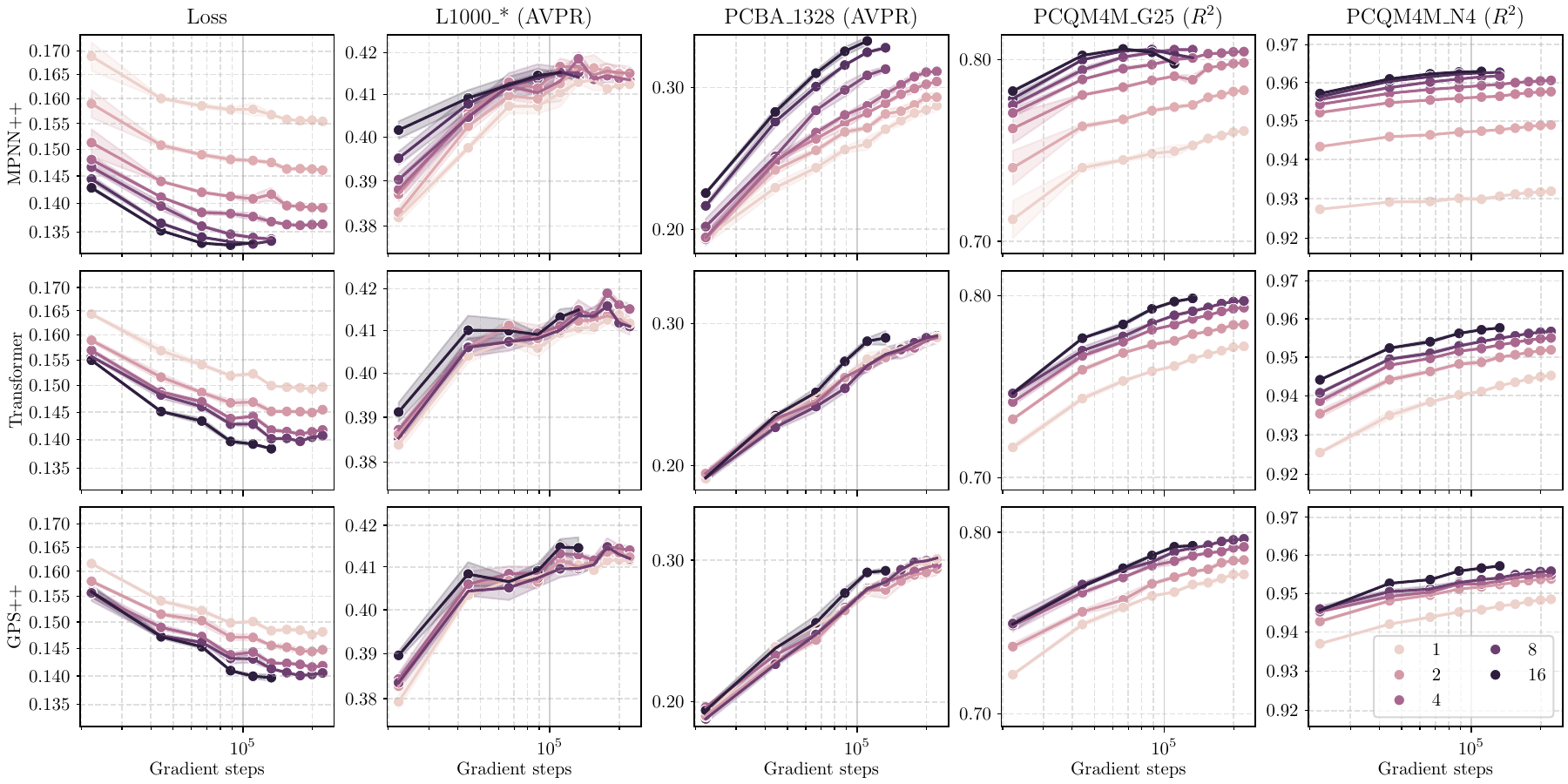}
    \caption{Model performance on the test set throughout training for MPNN++, Transformer, and GPS++ architectures with depth scaling. Different colors represent models with varying number of network layers.}
\end{figure}

\begin{figure}[H]
    \includegraphics[width=\textwidth]{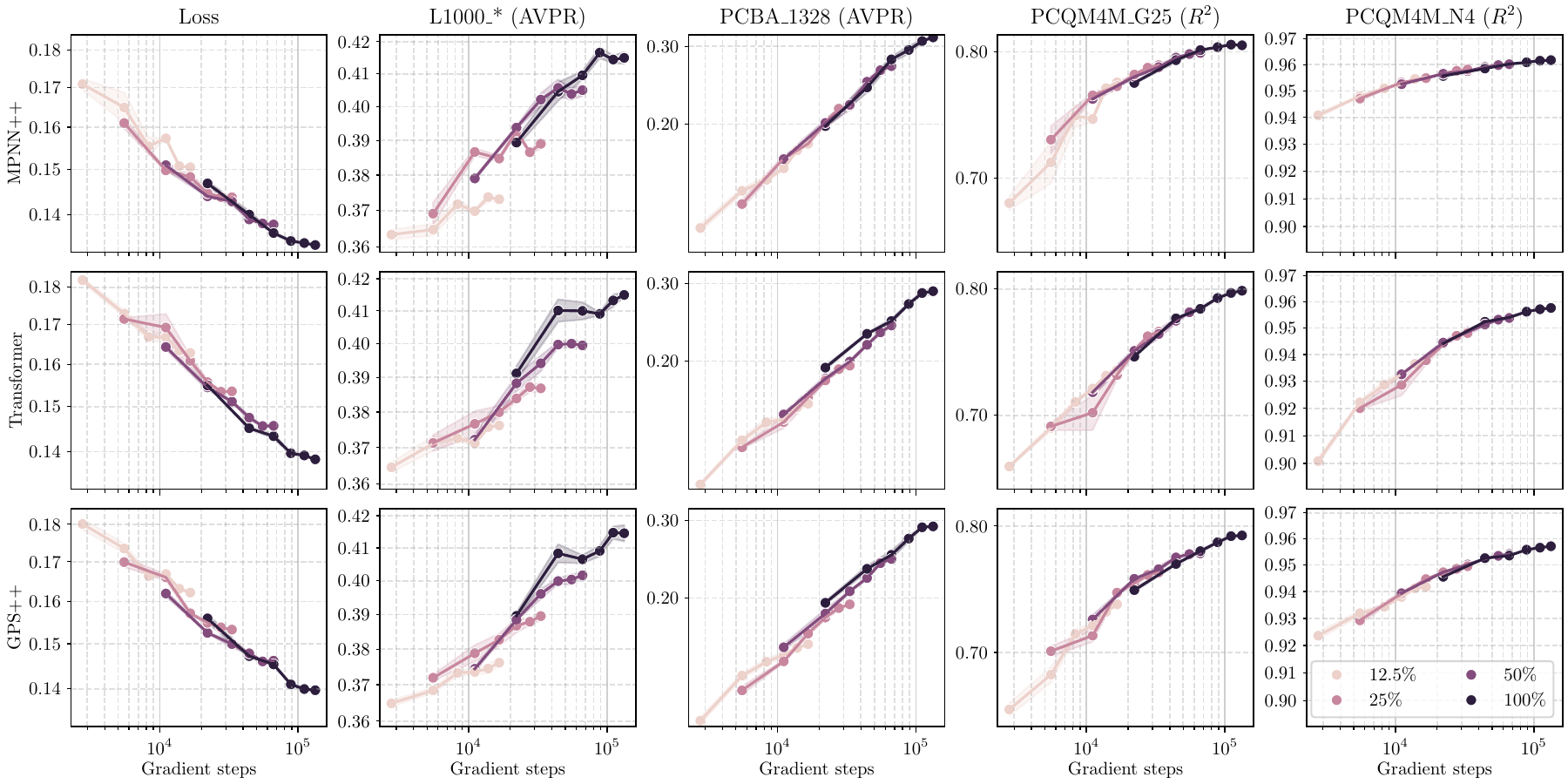}
    \caption{Model performance on the test set throughout training for MPNN++, Transformer, and GPS++ architectures with molecule scaling. Different colors represent models with varying fraction of molecules used for training.}
\end{figure}

\section{Additional Results on Downstream Tasks}
\label{sec:additional_downstream_results}

\begin{landscape}
\subsection{Width Scaling}
\label{subsec:app:width_scaling}

\begin{figure}[H]
    \vspace{-0.25cm}
    \centering
    \includegraphics[width=0.55\textwidth]{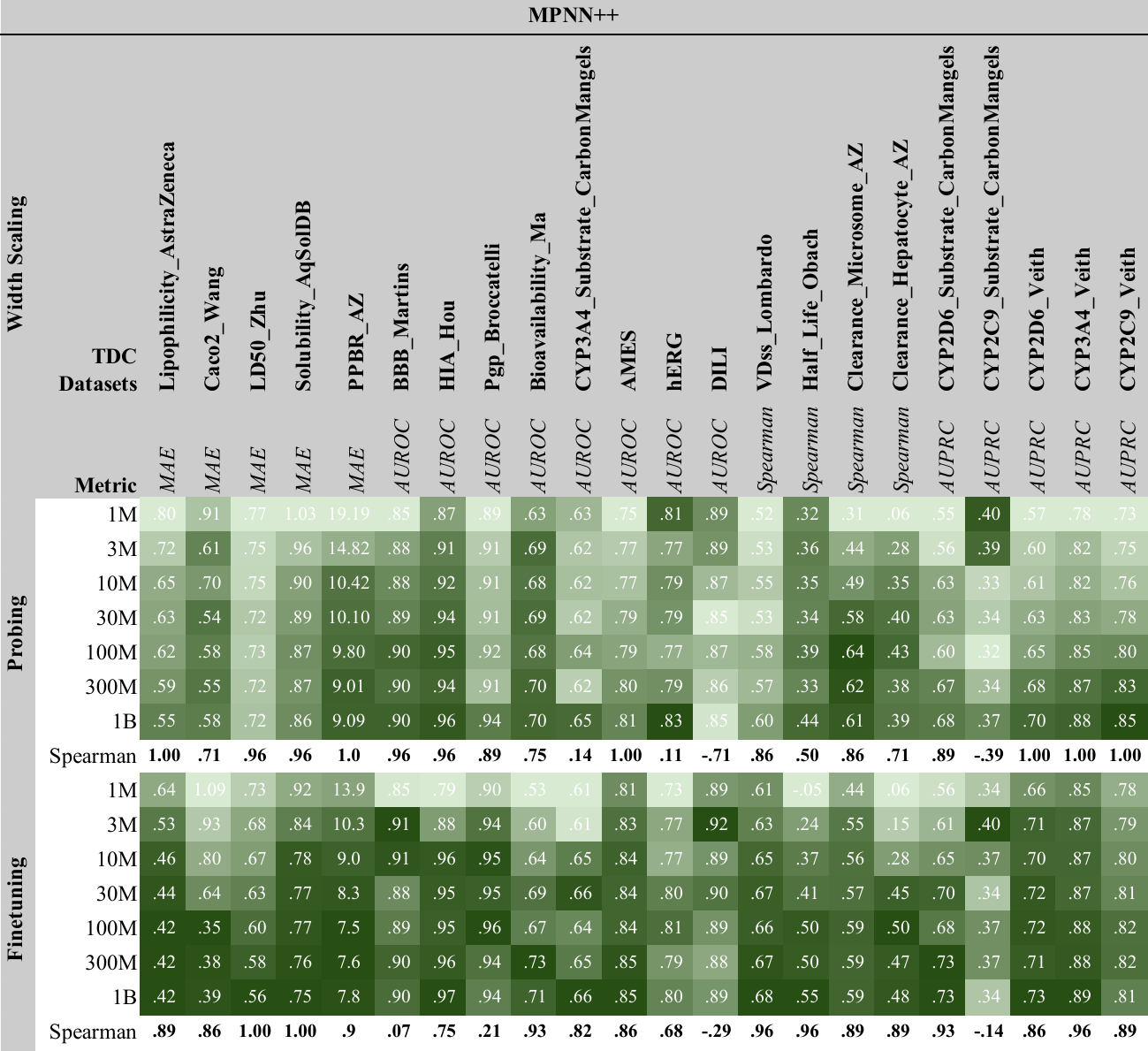}
    \raisebox{48pt}{
    \includegraphics[width=0.53\textwidth]{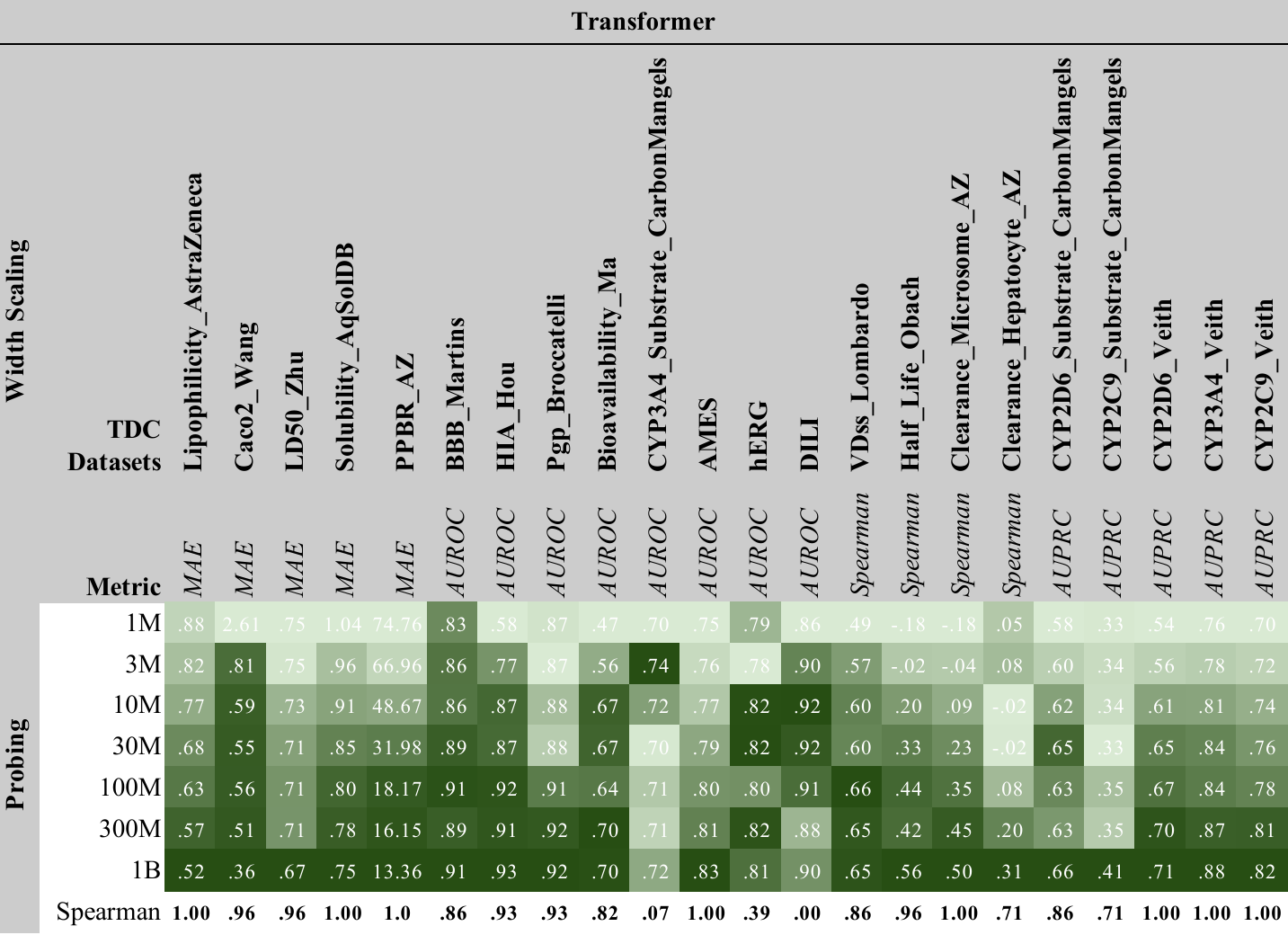}}
    \raisebox{50pt}{
    \includegraphics[width=0.53\textwidth]{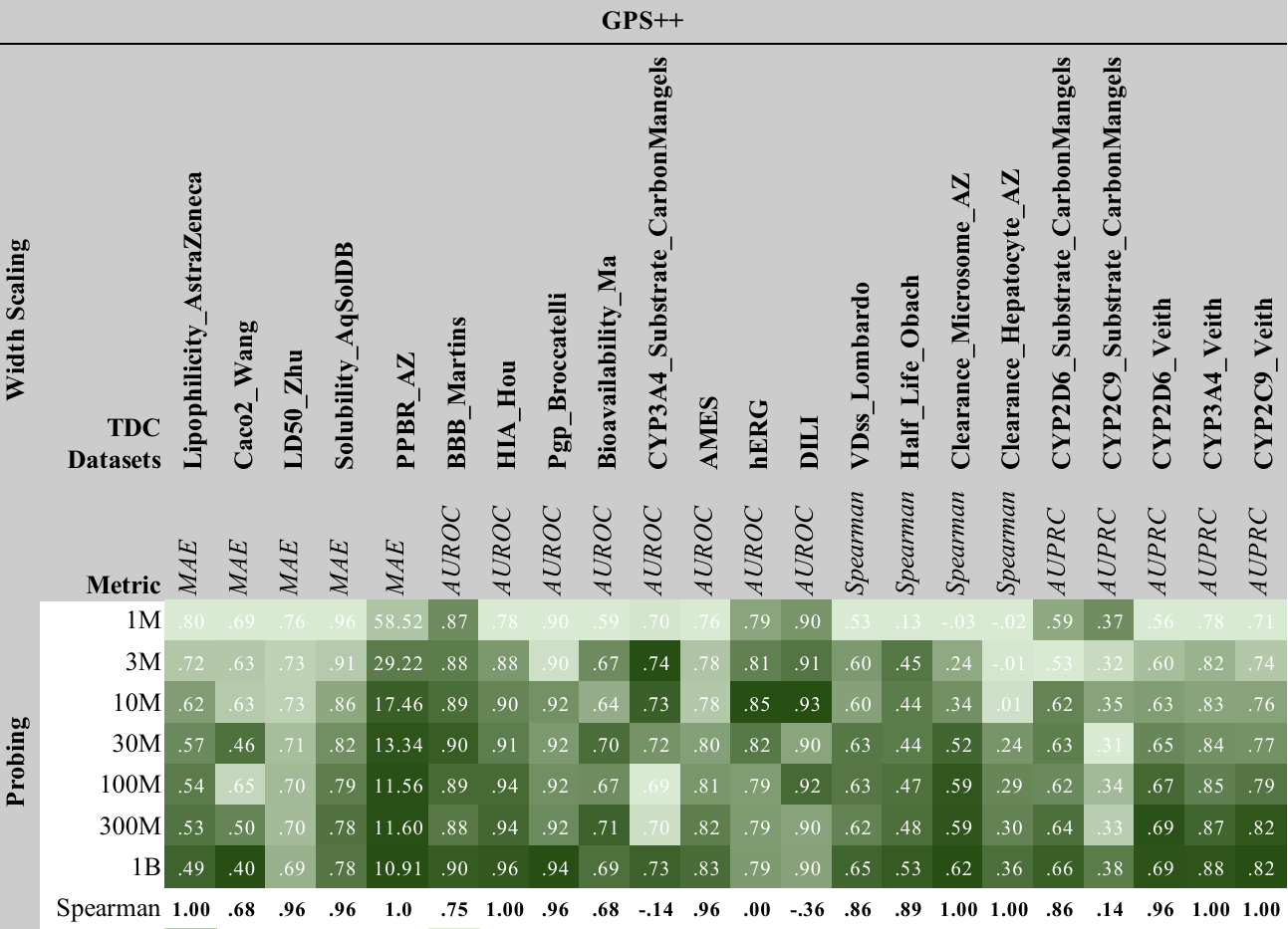}}
    \caption{\textbf{Width Scaling:} Comparison of probing and finetuning for MPNN++ \textbf{(left)}, Transformer \textbf{(center)}, and hybrid GPS++ \textbf{(right)} across different model sizes on the \underline{TDC} benchmark. \textbf{\color{dark2green}{Darker green}} shades denote higher/desirable metric values. Average Spearman correlations for MPNN++, Transformer and GPS++ models show improving scaling behavior with increasing number of parameters across the TDC benchmark. The average Spearman correlation between width and performance for probing is 0.69, 0.82 and 0.73, respectively, and 0.72 when finetuning MPNN++, effectively showing that model size plays an important role in predictive performance.}
    \label{fig:TDC_width}
    \vspace{-0.25cm}
\end{figure}
\end{landscape}

\begin{landscape}
\subsection{Depth Scaling}
\label{subsec:app:depth_scaling}

\begin{figure}[htbp]
    \centering
    \includegraphics[width=0.55\textwidth]{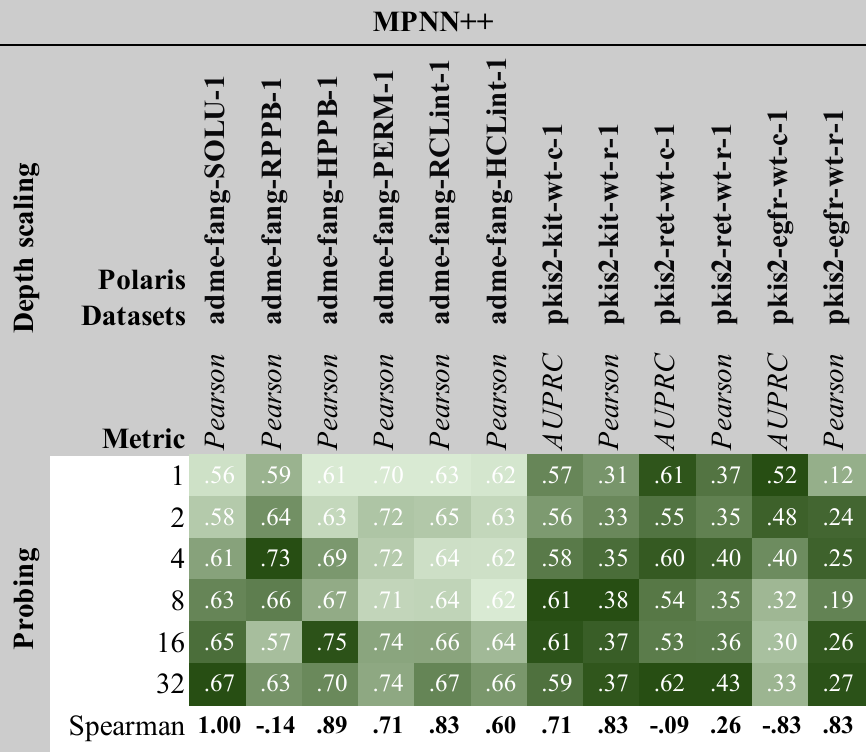}
    \raisebox{20pt}{
    \includegraphics[width=0.52\textwidth]{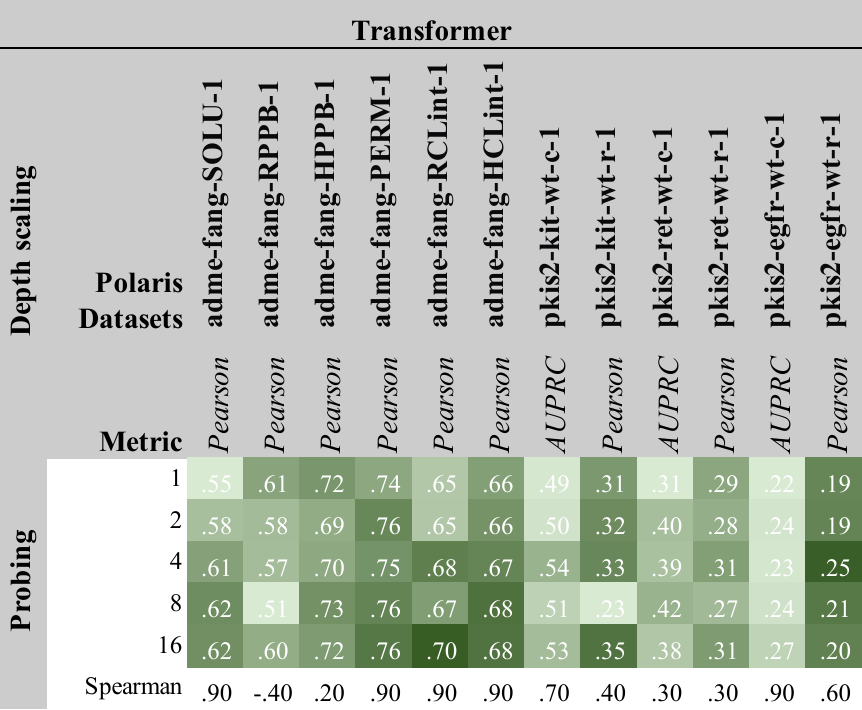}
    }
    \raisebox{20pt}{
    \includegraphics[width=0.52\textwidth]{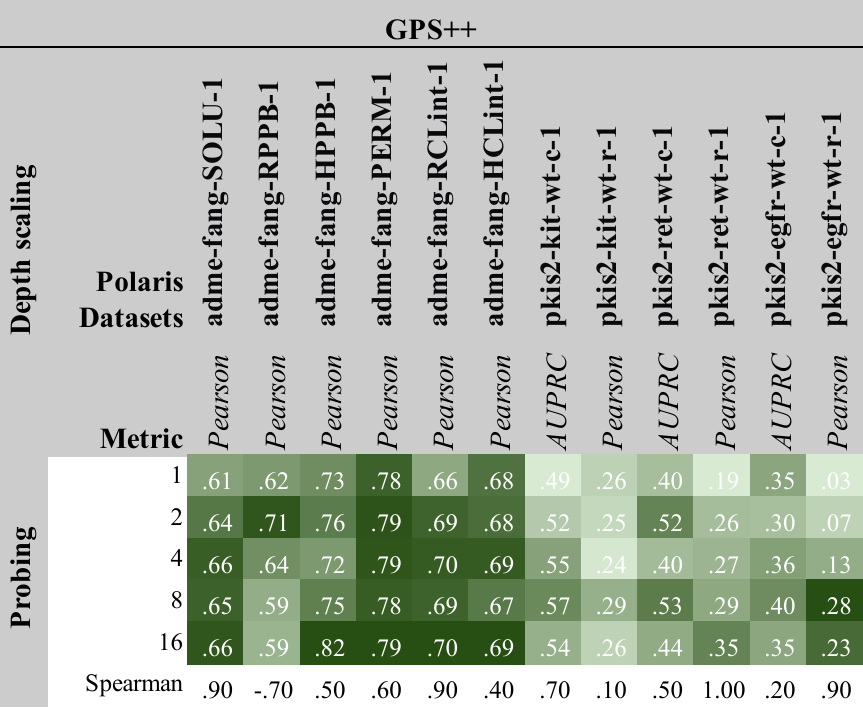}
    }
    \caption{\textbf{Depth Scaling:} Comparison of probing for different model depths on the \underline{Polaris} benchmark with MPNN++ \textbf{(left)}, Transformer \textbf{(center)}, and hybrid GPS++ \textbf{(right)}. \textbf{\color{dark2green}{Darker green}} shades denote higher/desirable metric values. Average Spearman correlation between depth and performance is \textit{$0.47$}, \textit{$0.55$}, and \textit{$0.50$}, respectively. Probing shows positive scaling trend with increasing depth across the Polaris benchmark.}
    \label{fig:polaris_depth_MPNN}
\end{figure}



\begin{figure}[htbp]
    \centering
    \includegraphics[width=0.55\textwidth]{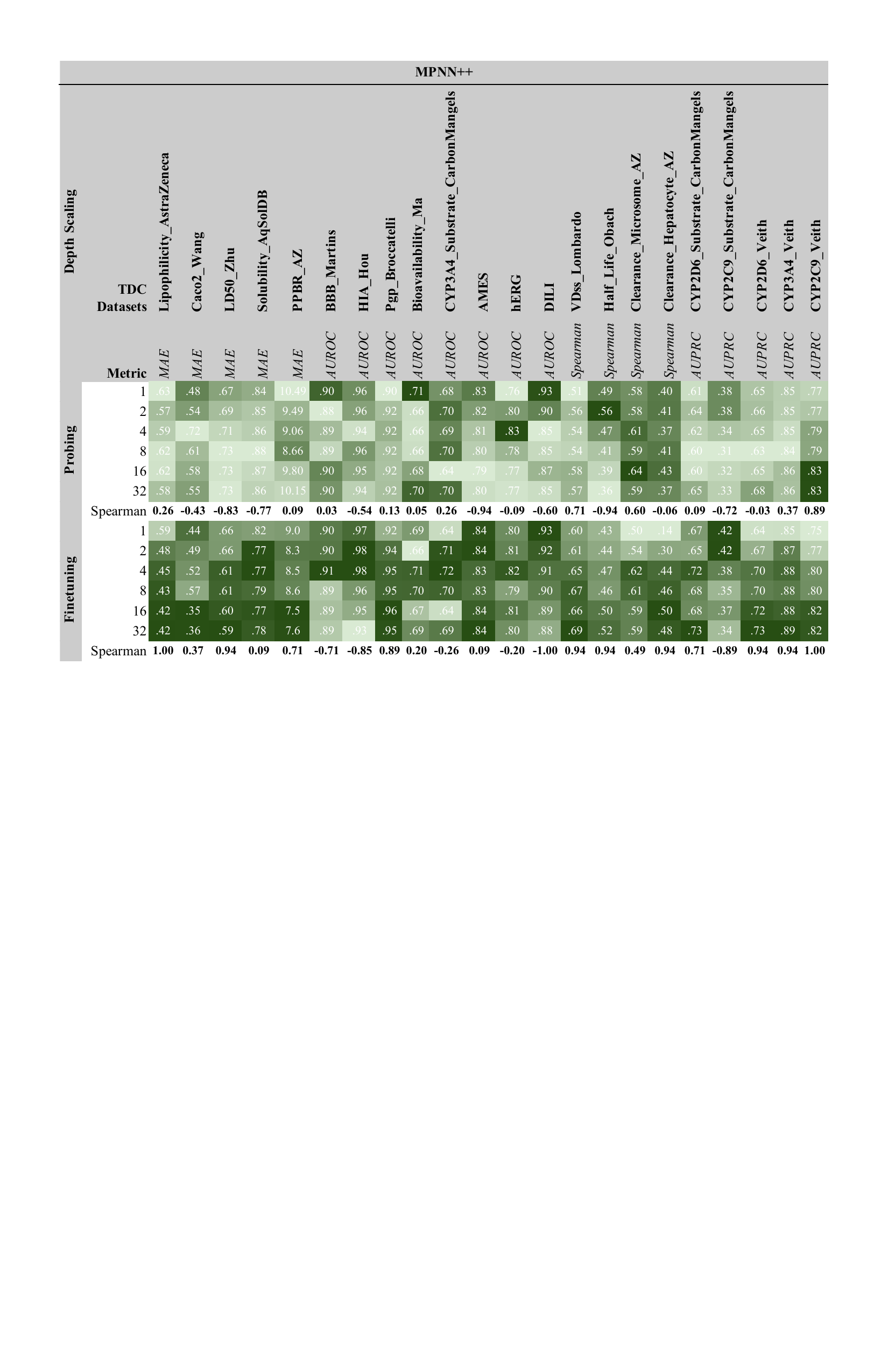}
    \raisebox{47pt}{
    \includegraphics[width=0.52\textwidth]{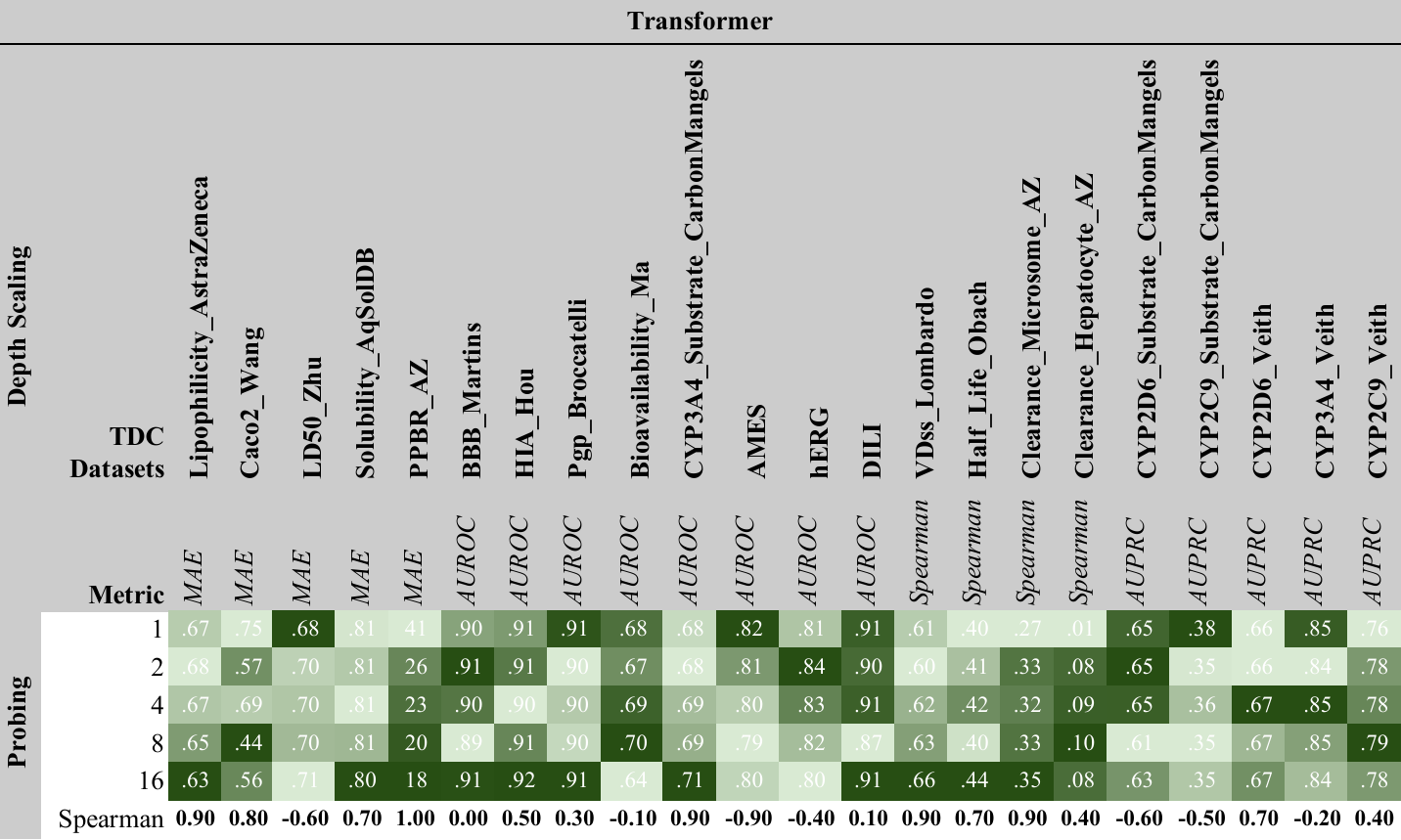}
    }
    \raisebox{52pt}{
    \includegraphics[width=0.52\textwidth]{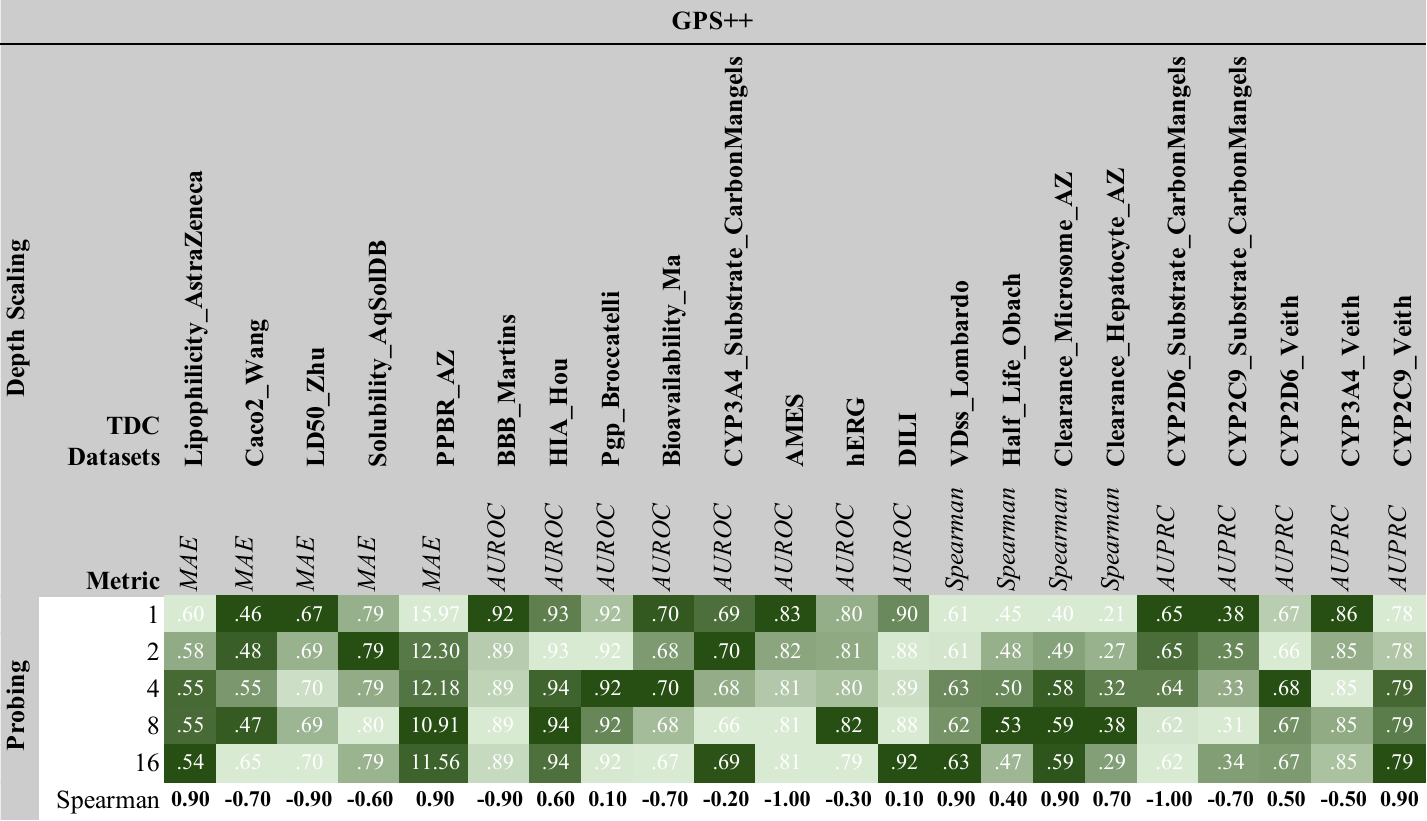}
    }
    \caption{\textbf{Depth Scaling:} Comparison of probing and finetuning for MPNN++ \textbf{(left)}, Transformer \textbf{(center)}, and GPS++ \textbf{(right)} models across different model depths on the \underline{TDC} benchmark. \textbf{\color{dark2green}{Darker green}} shades denote higher/desirable metric values. Average Spearman correlation between depth and probing performance is -0.11, 0.27 and -0.03, respectively, and 0.33 for finetuned MPNN++. While performance mostly increases with depths up to 8, larger depths often show signs of saturation.}
    \label{fig:TDC_depth_MPNN}
\end{figure}

\end{landscape}


\begin{landscape}

\subsection{Molecule Scaling}
\label{subsec:app:molecule_scaling}



\begin{figure}[htbp]
    \vspace{-0.25cm}
    \centering
    \includegraphics[width=0.55\textwidth]{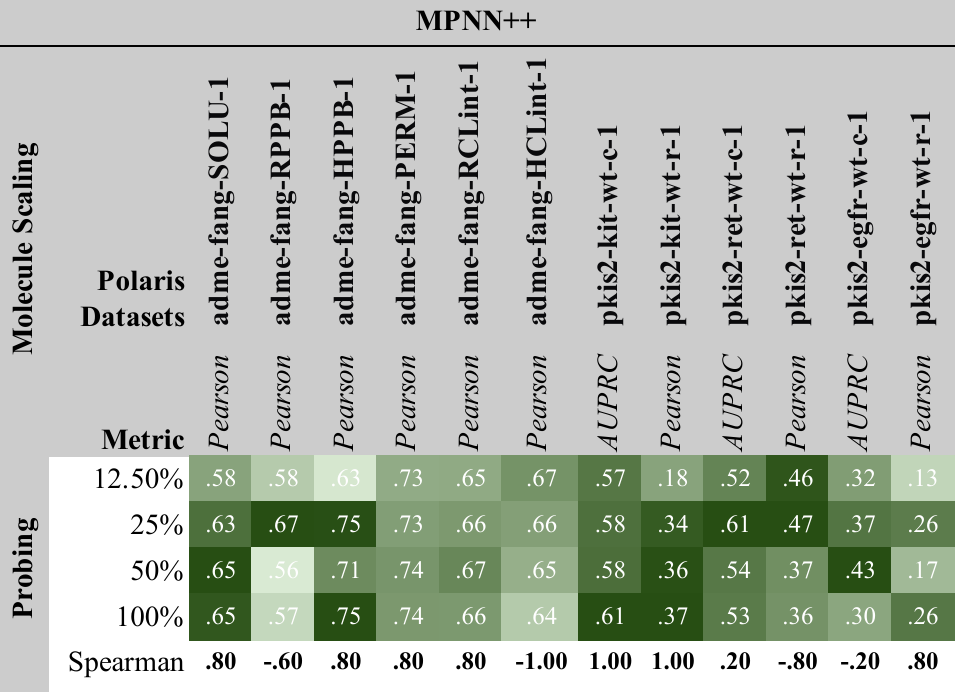}
    \raisebox{10pt}{
    \includegraphics[width=0.52\textwidth]{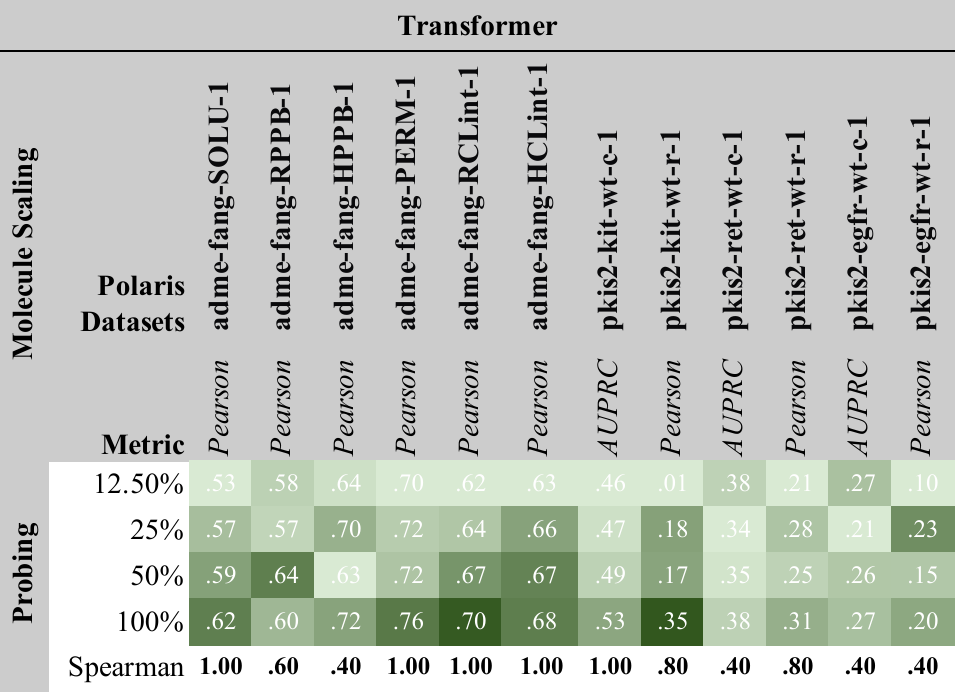}
    }
    \raisebox{10pt}{
    \includegraphics[width=0.52\textwidth]{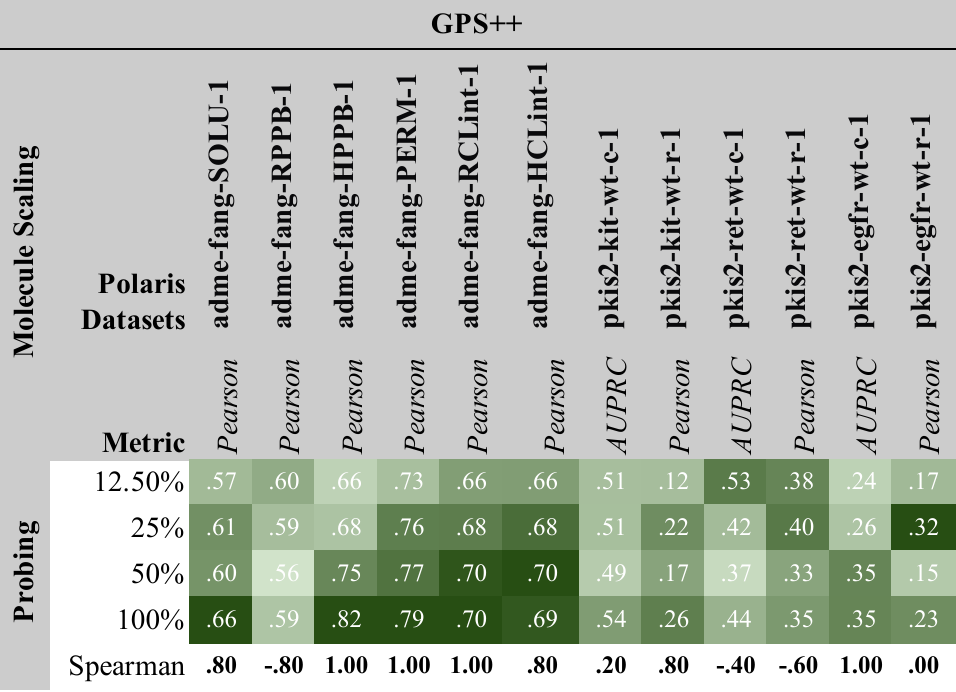}
    }
    \caption{\textbf{Molecule Scaling:} Scaling behavior of probed 100M parameter models across different dataset molecule fractions on the \underline{Polaris} benchmark with MPNN++ \textbf{(left)}, Transformer \textbf{(center)}, and hybrid GPS++ \textbf{(right)}. \textbf{\color{dark2green}{Darker green}} shades denote higher/desirable metric values. Average Spearman correlation between molecule fraction and probing performance is 0.30, 0.73, and 0.40, respectively. Models show consistent improvement in performance with the increasing size of datasets.}
    \label{fig:polaris_molecule}
    \vspace{-0.25cm}
\end{figure}



\begin{figure}[htbp]
    \vspace{-0.25cm}
    \centering
    \includegraphics[width=0.55\textwidth]{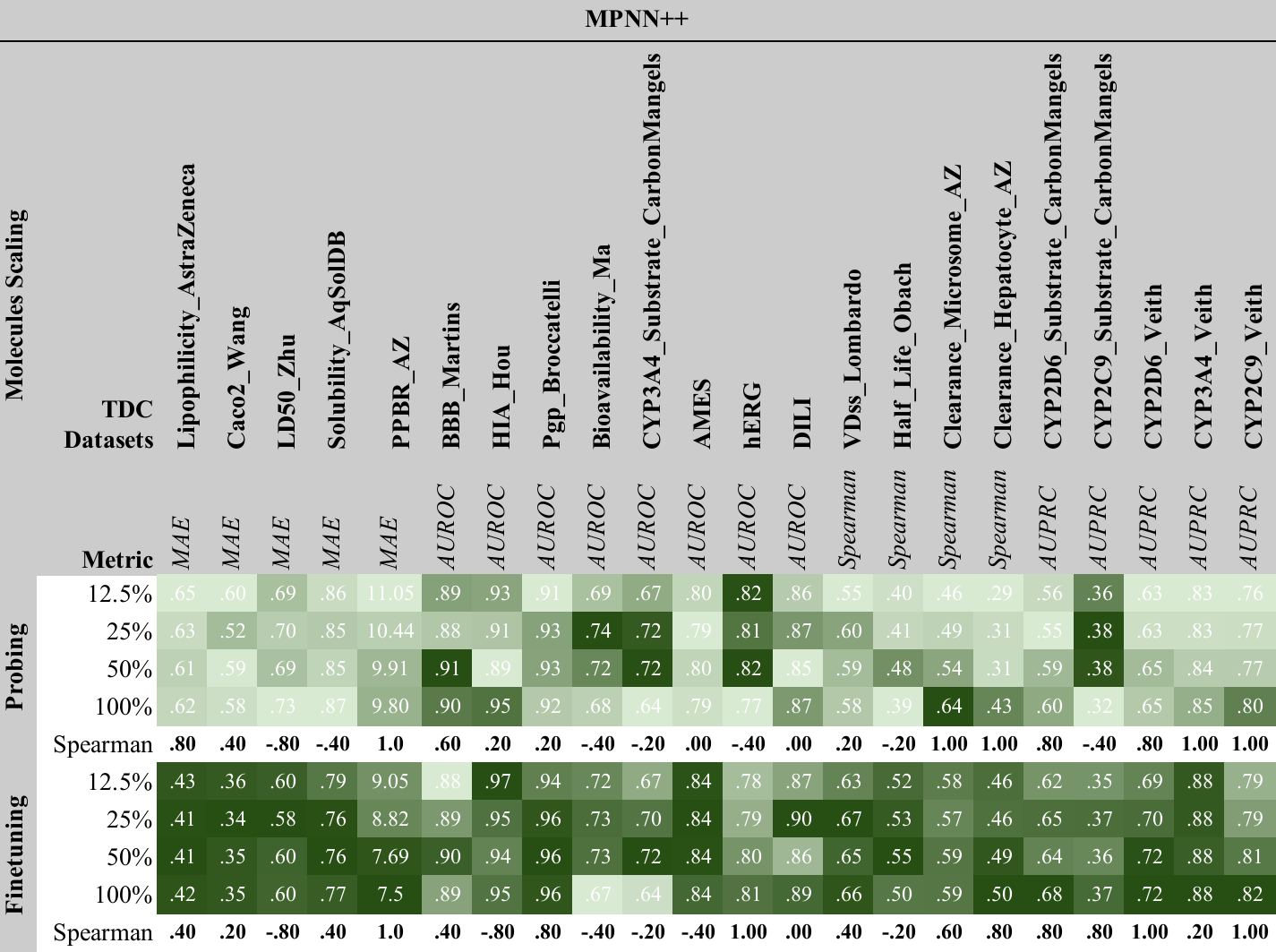}
    \raisebox{38pt}{
    \includegraphics[width=0.53\textwidth]{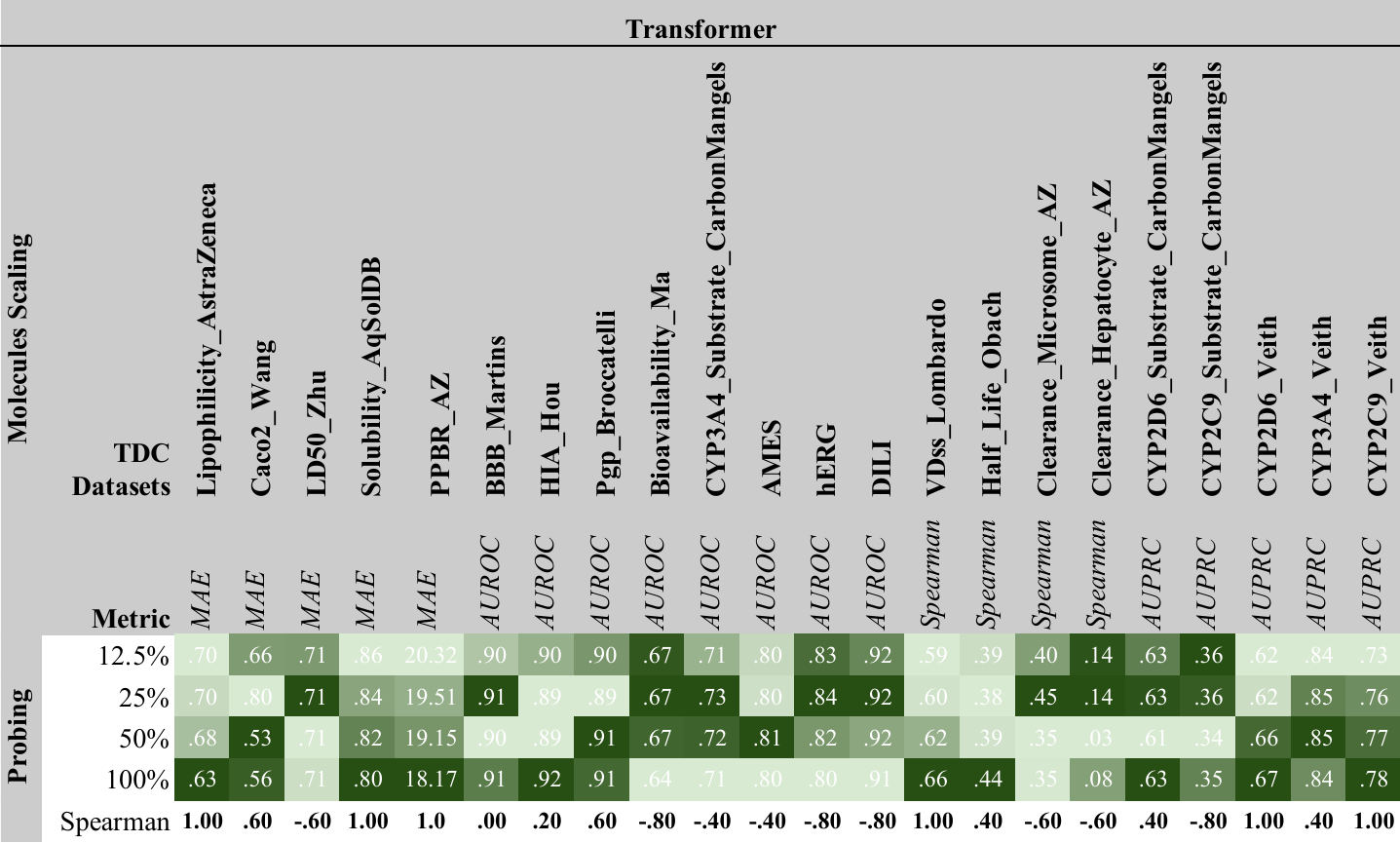}}
    \raisebox{38pt}{
    \includegraphics[width=0.53\textwidth]{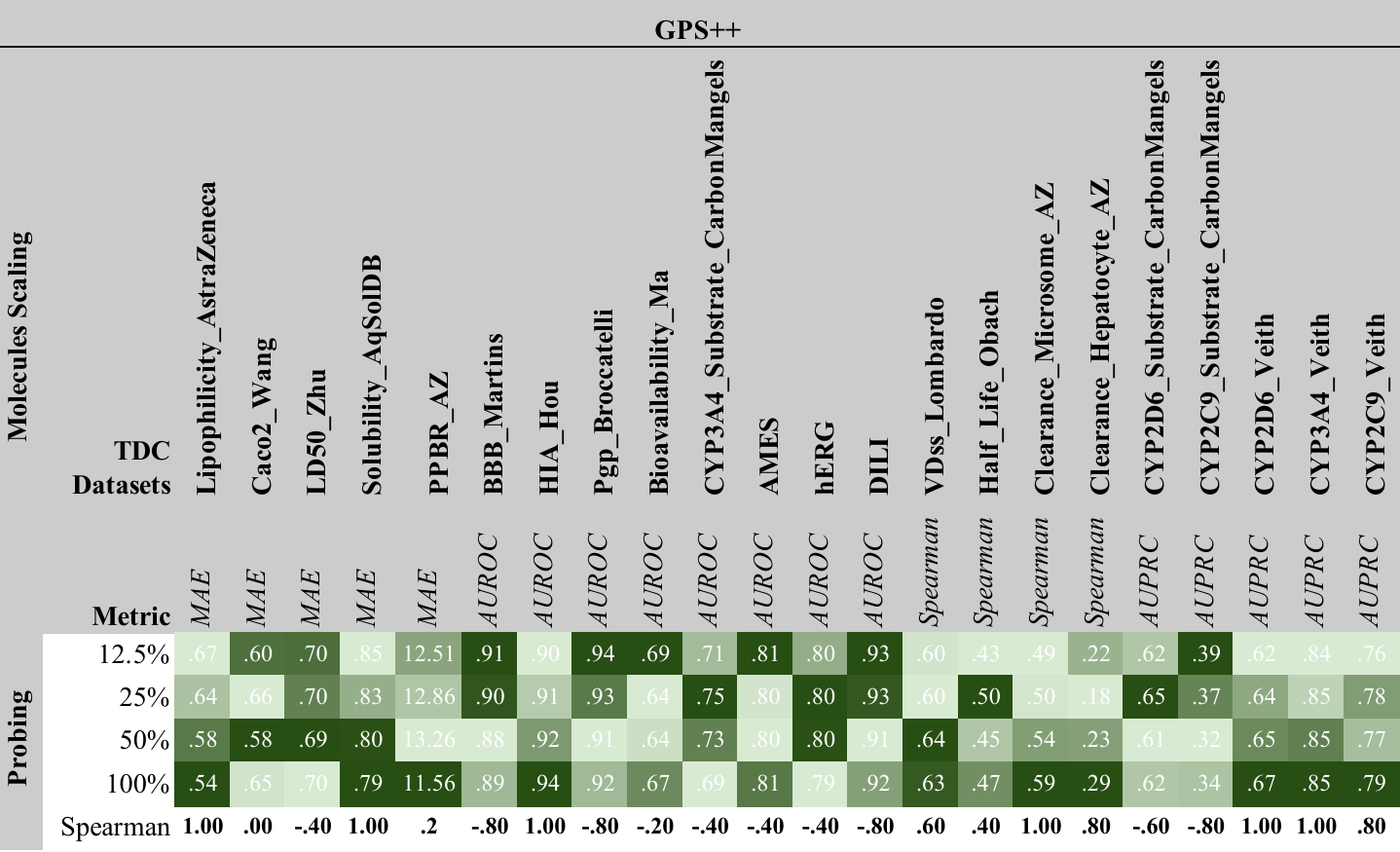}}
    \caption{\textbf{Molecule Scaling:} Comparison of probing and finetuning for MPNN++ \textbf{(left)}, Transformer \textbf{(center)} and GPS++ \textbf{(right)} models across different dataset sizes on the \underline{TDC} benchmark. \textbf{\color{dark2green}{Darker green}} shades denote higher/desirable metric values. Average Spearman correlation between molecule fraction and probing performance is 0.28, 0.13 and 0.15, respectively, and 0.32 for finetuned MPNN++. Finetuned models scale better when compared to probed models. However, increasing the size of finetuning datasets leads to minor improvements beyond the 50\% dataset size fraction.}
    \label{fig:TDC_molecule_MPNN}
    \vspace{-0.25cm}
\end{figure}

\end{landscape}


\subsection{Label Scaling}
\label{subsec:app:label_scaling}


\begin{figure}[H]
    \centering
    \includegraphics[width=0.7\textwidth]{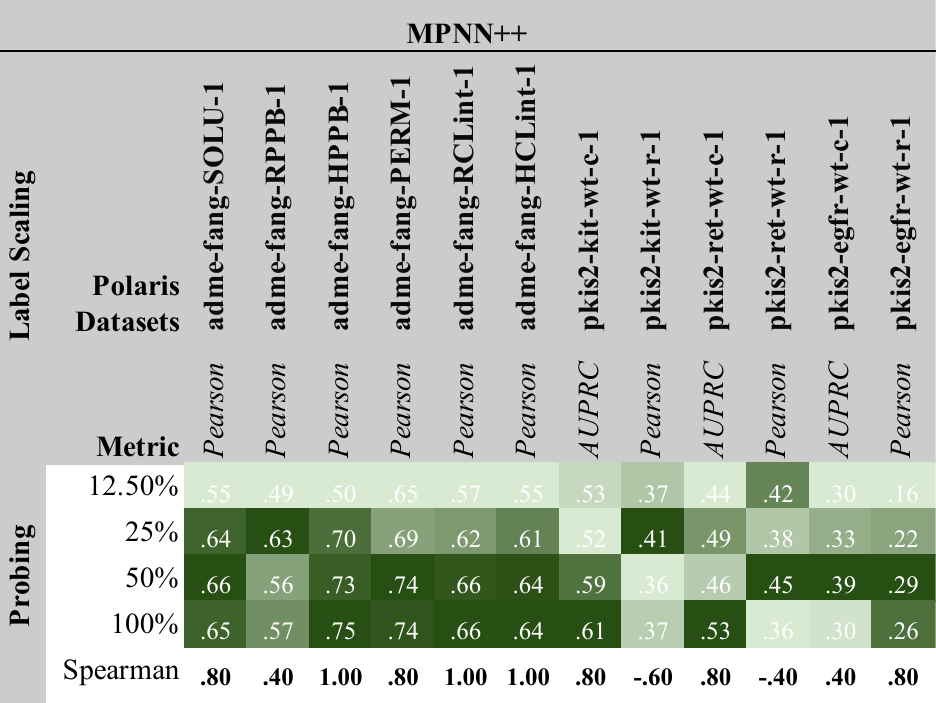}
    \caption{\textbf{Label Scaling:} Performance of MPNN++ probed models across different label fractions in the \underline{Polaris} benchmark. \textbf{\color{dark2green}{Darker green}} shades denote higher/desirable metric values. Average Spearman correlation between label fraction and probing performance is 0.57.}
    \label{fig:polaris_label_MPNN}
\end{figure}


\begin{figure}[H]
    \vspace{-0.25cm}
    \centering
    \includegraphics[width=0.8\textwidth]{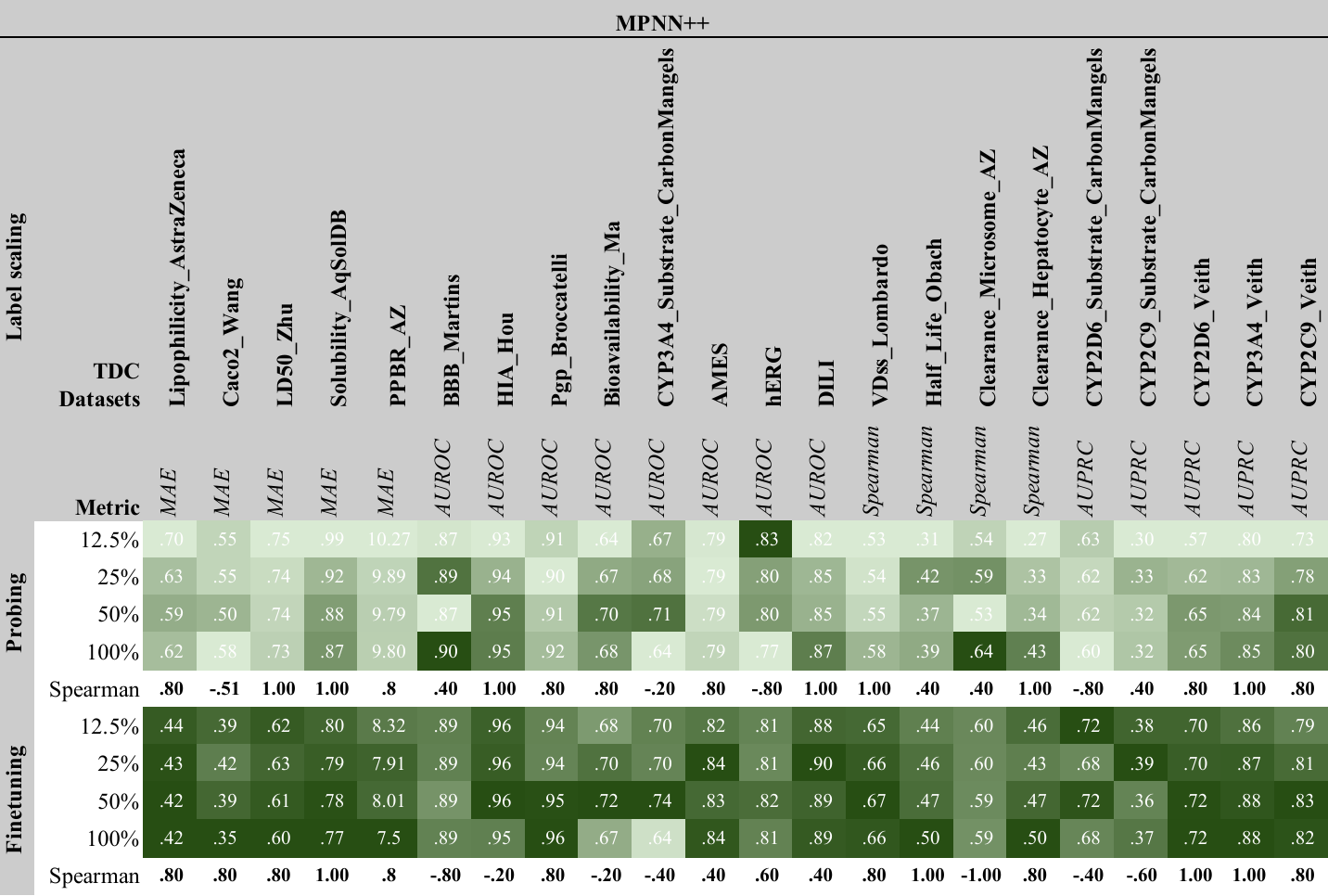}
    \caption{\textbf{Label Scaling:} Comparison of MPNN++ probing and finetuning across different label fractions on the \underline{TDC} benchmark. \textbf{\color{dark2green}{Darker green}} shades denote higher/desirable metric values. Average Spearman correlation between label fraction and performance is 0.54 for probed MPNN++ and 0.37 for finetuned MPNN++. Finetuned models scale better when compared to probed models. Increasing label fractions do not deteriorate model performance.}
    \label{fig:TDC_label_MPNN}
    \vspace{-0.25cm}
\end{figure}


\subsection{Dataset Ablation}
\label{subsec:app:dataset_ablation}


\begin{figure}[H]
    \centering
    \includegraphics[width=0.7\textwidth]{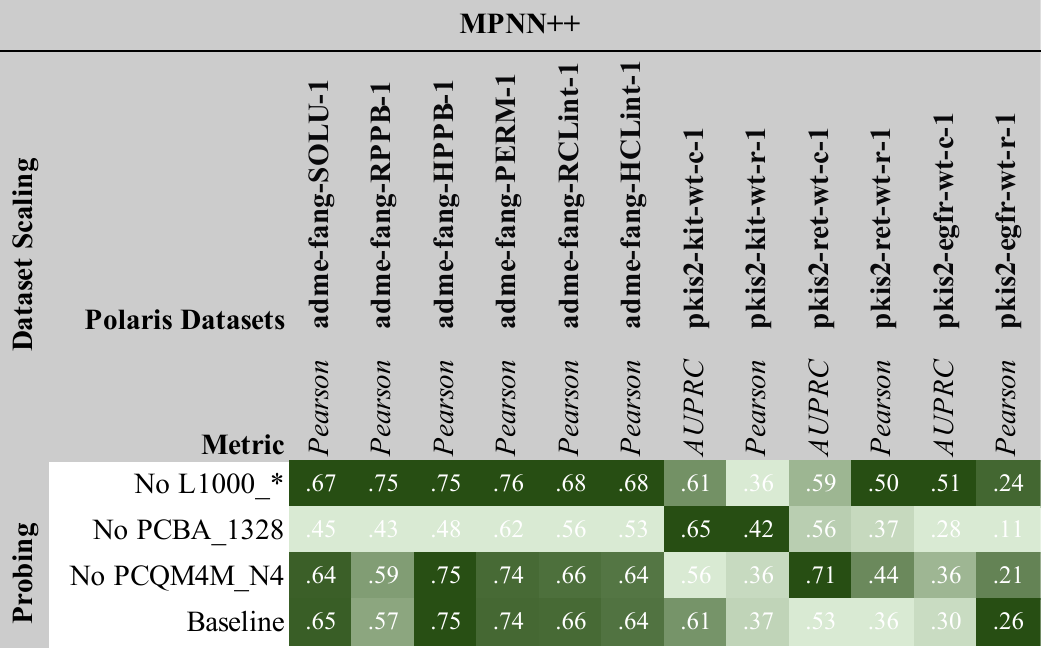}
    \caption{\textbf{Dataset Ablation:} Comparison of probed 100M parameter MPNN++ models on the \underline{Polaris} benchmark tasks (in columns) after pretrained \textbf{without} certain pretraining datasets (in rows). \textbf{\color{dark2green}{Darker green}} shades denote higher/desirable metric values. We observe that removing PCBA\_1328 significantly hurts downstream performance across almost all tasks, while removing the L1000 leads to noticeable improvements on most tasks.}
    \label{fig:Polaris_dataset_MPNN}
\end{figure}


\begin{figure}[H]
    \vspace{-0.25cm}
    \centering
    \includegraphics[width=0.9\textwidth]{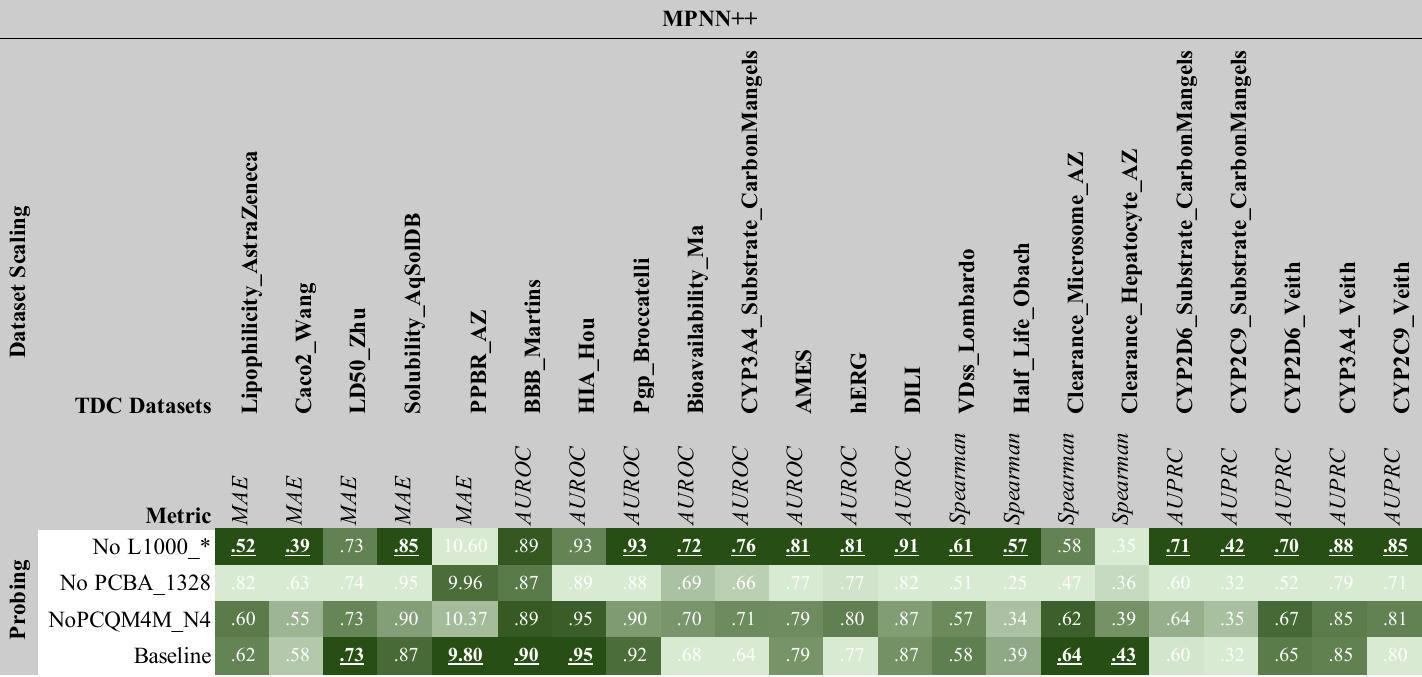}
    \caption{\textbf{Dataset Ablation:} Performance of probed 100M parameter MPNN++ models on \underline{TDC} benchmark tasks (in columns) after pretrained \textbf{without} certain datasets (in rows). \textbf{\color{dark2green}{Darker green}} shades denote higher/desirable metric values.  We observe that removing PCBA\_1328 significantly hurts downstream performance across almost all tasks, while removing the L1000 leads to noticeable improvements on most tasks.}
    \label{fig:TDC_dataset_MPNN}
    \vspace{-0.25cm}
\end{figure}

\subsection{Task-Head Ablation}
\label{subsec:app:task_head_ablation}

\begin{figure}[H]
    \vspace{-0.25cm}
    \centering
    \includegraphics[width=0.9\textwidth]{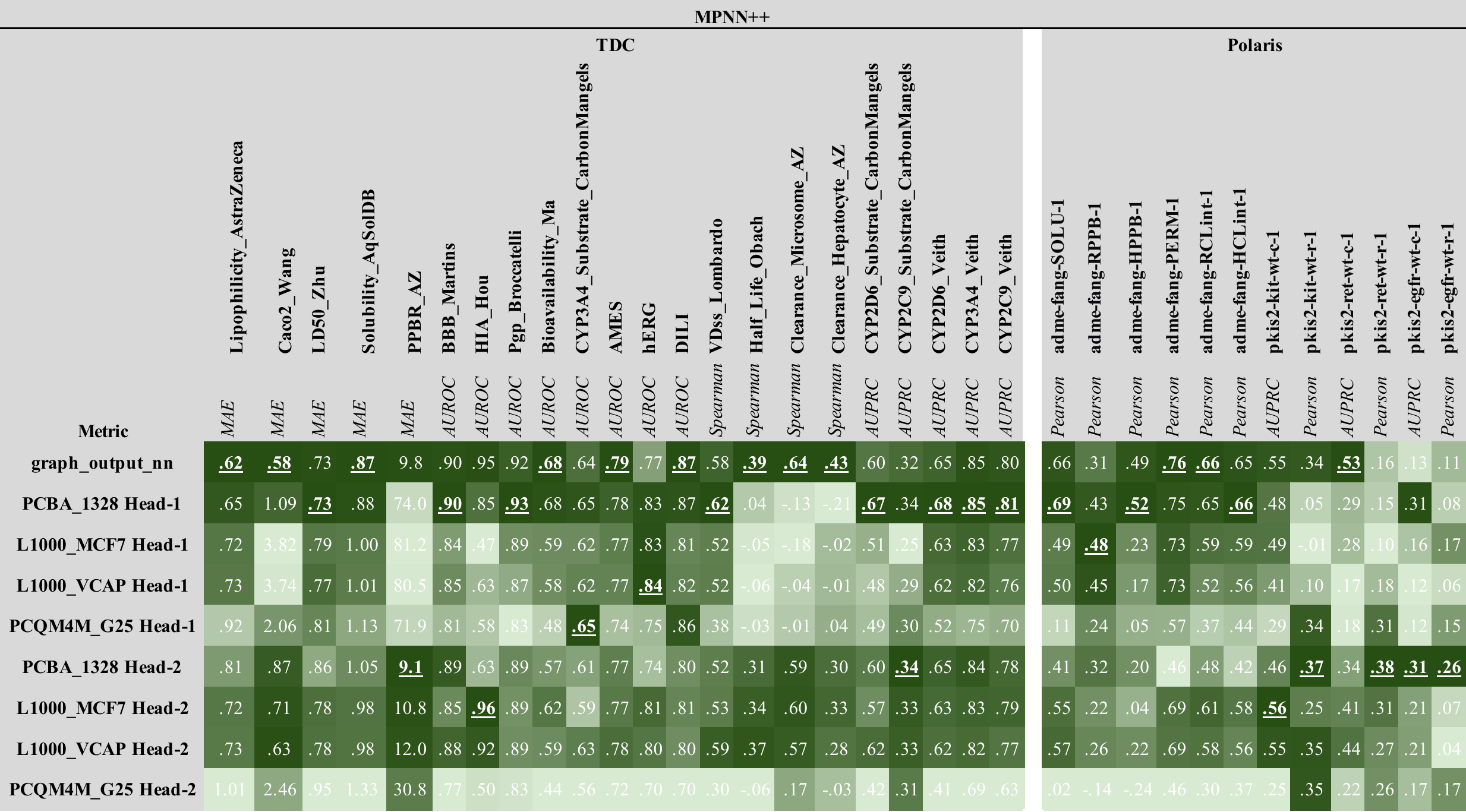}
    \caption{\textbf{Task-Head Ablation:} Performance of probed 100M parameter MPNN++ models on \underline{TDC} benchmark (left) and \underline{Polaris} benchmark (right), with the tasks in columns, when probing from different task heads (in rows). \textbf{\color{dark2green}{Darker green}} shades denote higher/desirable metric values, and \textbf{\underline{bold/underline}} indicates the best value in a given column. We observe that the \textit{graph\_output\_nn} (e.g. the hidden representation that is fed to all task-heads) is overall the best choice for probing. We hypothesize this is because it captures and compresses the combined information from the various pretraining tasks. The PCBA\_1328 task-head is also a good choice due to its proximity to the considered downstream tasks. The PCQM4M\_G25 task-head is the least useful as the tasks are fundamentally different from the downstream tasks.}
    \label{fig:TDC_POLARIS_task_head_mpnn}
    \vspace{-0.25cm}
\end{figure}


\subsection{The TDC Benchmark -- Data Leakage}
\label{subsec:app:TDC-leakage}

Considering that the pretraining dataset is supervised, it is important to consider data-leakage as a source of experimental error. This is especially the case for the PCBA\_1328 dataset.

PCBA\_1328 contains only classification assays with more than 6000 molecules each, which automatically disqualifies most of TDC and all of Polaris. The TDC datasets remaining after this constraint are the 3 CYP*\_Veith datasets, and the AMES dataset. 
The AMES dataset is not present in PubChem \cite{pubchem2023}, excluding it from the list of potential leaks. 
Regarding the 3 CYP*\_Veith datasets, they represent inibition assays against recombinant enzymes \cite{veith2009comprehensive}. The three assays from TDC, and two others from the paper, are all present and aggregated under assayID-1851. Therefore, whenever a molecule is active against any of the enzyme, the value is $1$, otherwise it is $0$. Therefore, there is a minor leak, although the datasets are not identical. Further, no evidence of leak was observed in terms of abnormally high performance of the model on these assays, which is expected considering that the model is learning more than 3000 labels simultaneously.

\newpage
\section{Comparison of MolGPS to Unsupervised Methods}\label{subsec:app:unsupervised}

\begin{table}[t!]
    \centering
    \footnotesize
    \caption{Comparison of MolGPS variants to the self-supervised GraphMVP model on the MoleculeNet dataset (test AUROC). Note that we only consider the datasets that were not part of our pretraining.}
    \label{tab:unsupervised-comp-1}
    \vspace{10pt}
    \label{tab:your-table}
    \begin{tabular}{lcccc}
        \toprule
        Method                      & BACE & BBBP & Clintox & Sider \\
        \midrule
        GraphMVP                    & 0.812 & 0.724 & 0.775 & 0.639 \\
        1B MolGPS w/o Phenomics   & 0.806	& \bf{0.802}	& 0.797	& 0.649 \\
        1B MolGP  & \bf{0.828}	    & 0.8	& \underline{\bf{0.809}}	    & \underline{\bf{0.67}} \\
        3B MolGPS                   & \underline{\bf{0.832}}	& \underline{\bf{0.809}}	    & \bf{0.807}	& \bf{0.666} \\
        \bottomrule
    \end{tabular}
\end{table}

Compared to unsupervised pretraining approaches, it is important to remark that the scale of pretraining data is not directly comparable to that of our supervised pretraining approach, i.e, billions of molecules in some cases. We note that previous works in the GNN literature~\citep{chen2024uncovering} found interesting scaling trends despite scaling to even less than the 5M graphs that we have aggregated in the LargeMix dataset.

In our supervised context, the data scale does not only depend on the number of molecules seen during pretraining. Instead, each molecule should be considered in conjunction with the set of pretraining labels. PCBA\_1328 considers more than 1k different labels per molecule (albeit with high sparsity) and PCQM4M comes with 25 graph-level tasks and 4 node-level tasks (for each node of the ~4M molecules). We point to our label scaling study that shows the impact of reduced data diversity by removing labels (e.g., Figure~\ref{fig:pretraining_one}) for convincing evidence for this trend. The performance gains from incorporating Phenomics data into the pretraining data mix also support this claim, where the addition of ~500k molecules with ~6k highly informative labels per graph (Figure~\ref{fig:sota_aggregated}) leads to strong performance gains.

Here, we provide further comparison to unsupervised pretraining approaches (Table~\ref{tab:unsupervised-comp-1} and~\ref{tab:unsupervised-comp-2}). We observe that the unsupervised MolE model variants~\citep{mole} clearly underperform the standard MolE model that leverages both unsupervised and supervised pretraining (Table~\ref{tab:unsupervised-comp-2}), which is in turn outperformed by the listed MolGPS variants by large margins. Even our smaller 100M MPNN++ models (that are of comparable parametric size to MolE) both outperform the self-supervised variants (see Figure~\ref{fig:sota_aggregated}).

In Table~\ref{tab:unsupervised-comp-1}, we compare to the self-supervised GraphMVP~\citep{graphmvp} model on MoleculeNet. MolGPS without Phenomics pretraining outperforms GraphMVP on 3/4 tasks, while the model variants that were pretrained with additional Phenomics data outperform across all tasks.

\begin{table}[h!]
    \centering
    \footnotesize
    \caption{Comparison of MolGPS variants to the self-supervised variants of MolE and the standard MolE (supervised + self-supervised pretraining) on TDC benchmark collection. Supervised pretraining significantly improves MolE performance as seen in the normalized performance (left-most column). MolGPS models outperform all MolE models by a large margin, exhibiting the best performance in all but two tasks.}
    \label{tab:unsupervised-comp-2}
    \vspace{10pt}
    \scalebox{0.45}{
        \begin{tabular}{l|c|cccccccccccccccccccccc}
        \toprule
        & &\rotatebox{90}{\textbf{Lipophilicity\_AstraZeneca}} &\rotatebox{90}{\textbf{Caco2\_Wang}} &\rotatebox{90}{\textbf{LD50\_Zhu}} &\rotatebox{90}{\textbf{Solubility\_AqSolDB}} &\rotatebox{90}{\textbf{PPBR\_AZ}} &\rotatebox{90}{\textbf{BBB\_Martins}} &\rotatebox{90}{\textbf{HIA\_Hou}} &\rotatebox{90}{\textbf{Pgp\_Broccatelli}} &\rotatebox{90}{\textbf{Bioavailability\_Ma}} &\rotatebox{90}{\textbf{CYP3A4\_Substrate\_CarbonMangels}} &\rotatebox{90}{\textbf{AMES}} &\rotatebox{90}{\textbf{hERG}} &\rotatebox{90}{\textbf{DILI}} &\rotatebox{90}{\textbf{VDss\_Lombardo}} &\rotatebox{90}{\textbf{Half\_Life\_Obach}} &\rotatebox{90}{\textbf{Clearance\_Microsome\_AZ}} &\rotatebox{90}{\textbf{Clearance\_Hepatocyte\_AZ}} &\rotatebox{90}{\textbf{CYP2D6\_Substrate\_CarbonMangels}} &\rotatebox{90}{\textbf{CYP2C9\_Substrate\_CarbonMangels}} &\rotatebox{90}{\textbf{CYP2D6\_Veith}} &\rotatebox{90}{\textbf{CYP3A4\_Veith}} &\rotatebox{90}{\textbf{CYP2C9\_Veith}} \\\midrule
        \bf{Method}  &\rotatebox{90}{Norm.} &\rotatebox{90}{MAE} &\rotatebox{90}{MAE} &\rotatebox{90}{MAE} &\rotatebox{90}{MAE} &\rotatebox{90}{MAE} &\rotatebox{90}{AUROC} &\rotatebox{90}{AUROC} &\rotatebox{90}{AUROC} &\rotatebox{90}{AUROC} &\rotatebox{90}{AUROC} &\rotatebox{90}{AUROC} &\rotatebox{90}{AUROC} &\rotatebox{90}{AUROC} &\rotatebox{90}{Spear.} &\rotatebox{90}{Spear.} &\rotatebox{90}{Spear.} &\rotatebox{90}{Spear.} &\rotatebox{90}{AUPRC} &\rotatebox{90}{AUPRC} &\rotatebox{90}{AUPRC} &\rotatebox{90}{AUPRC} &\rotatebox{90}{AUPRC} \\\midrule
        \textbf{MolE-FuncEnv (unsupervised)} &0.47 &0.46	&0.355	&0.597	&0.799	&8.57	&0.895	&0.951	&0.873	&0.638	&0.612	&0.831	&\underline{\bf{0.871}}	&0.89	&0.622	&0.579	&0.567	&0.373	&\underline{\bf{0.715}}	&0.411	&0.678	&0.857	&0.759 \\
        \textbf{MolE-AtomEnv (unsupvervised)} &0.477 &0.464	&0.471	&0.582	&0.81	&8.191	&0.895	&0.949	&0.871	&0.683	&0.633	&0.832	&0.844	&0.883	&0.596	&0.518	&0.531	&0.367	&0.706	&0.429	&0.665	&0.865	&0.773 \\
        \textbf{MolE} &0.728 &0.469	&0.31	&\bf{0.577}	&0.792	&8.073	&0.903	&0.963	&0.915	&0.654	&0.67	&0.813	&0.823	&0.883	&\bf{0.654}	&0.549	&0.607	&0.381	&0.699	&0.446	&0.682	&0.867	&0.801 \\
        \textbf{1B MolGPS w/o Phenomics} &1.222 &0.4	&0.347	&0.645	&0.714	&\underline{\bf{6.249}}	&0.922	&\underline{\bf{0.984}}	&0.941	&0.64	&\underline{\bf{0.681}}	&0.8389	&0.86	&\bf{0.937}	&\underline{\bf{0.655}}	&\underline{\bf{0.64}}	&\underline{\bf{0.659}}	&\bf{0.56}	&0.712	&\underline{\bf{0.483}}	&\bf{0.747}	&\underline{\bf{0.905}}	&\underline{\bf{0.871}} \\
        \textbf{1B MolGPS} &\bf{1.305} &\bf{0.391}	&\underline{\bf{0.288}}	&0.589	&\bf{0.706}	&6.497	&\bf{0.939}	&0.975	&\bf{0.947}	&\bf{0.686}	&\bf{0.681}	&\bf{0.85}	&\bf{0.868}	&0.933	&0.649	&\bf{0.632}	&\bf{0.649}	&0.527	&0.700	&\bf{0.474}	&0.741	&0.898	&0.832 \\
        \textbf{3B MolGPS} &\underline{\bf{1.404}} &\underline{\bf{0.386}}	&\bf{0.292}	&\underline{\bf{0.557}}	&\underline{\bf{0.679}}	&\bf{6.464}	&\underline{\bf{0.941}}	&\bf{0.98}	&\underline{\bf{0.948}}	&\underline{\bf{0.701}}	&0.68	&\underline{\bf{0.857}}	&0.864	&\underline{\bf{0.942}}	&0.649	&0.631	&0.633	&\underline{\bf{0.57}}	&\bf{0.713}	&0.464	&\underline{\bf{0.75}}	&\bf{0.9}	&\bf{0.838} \\
        \bottomrule
        \end{tabular}
    }
\end{table}

\newpage

\section{Scaling Law Details}\label{sec:app:scaling-laws}

We explore the power law fit which governs the scaling behavior of our GNNs using Equations~\ref{eq:one}~and~\ref{eq:two} to compute the power law constants and identify data and parameter requirements. For metrics with higher desirable values (such as AUROC or R2), fractions inside the exponents are reversed.


Table~\ref{tab:power-law:polaris} presents values of \textit{$\alpha$} (Equation~\ref{eq:one}) based on model parameters and tasks considered in our downstream task experiments on the Polaris benchmark (Figure~\ref{fig:polaris_width_MPNN}). We choose $|\theta_{c}| = 1B$ and fix our final training error values corresponding to this model’s performance. On average, we see $\alpha \approx 0.081$ for probing and $\alpha \approx 0.098$ for finetuning. These relationships hold across 6 orders of magnitude in $|\theta|$ indicating that finetuning behavior scales logarithmically with the number of trainable parameters. It is also worth noting that \citet{scalinglaws} obtain similar $\alpha \approx 0.076$ indicating that our power law fit lies within the same parameter budget.

Table~\ref{tab:power-law:data} compares values of \textit{$\beta$} (Equation~\ref{eq:two}) for pretraining our models as presented in Figure~\ref{fig:pretraining_one}. We choose $|\mathcal{D}_{c}| = 5M$ to be \textit{$100\%$} of molecular data. As observed, $\beta \approx 0.110$ for MPNN++, $\beta \approx 0.106$ for Transformer and $\beta \approx 0.106$ for GPS++ for the overall test error. This holds for datasets up to 5M molecules. Notably, all $\beta > 0$ indicate that, given a computational budget, larger GNNs scale favorably. We again note that \citet{scalinglaws} have similar power law fits of $\beta \approx 0.095$ albeit with a $10^{13}$ token corpus. This allows us to draw two conclusions. Firstly, GNNs continue to demonstrate optimal scaling with limited datasets. Secondly, our 5M molecules (10M node features) dataset is sufficient to demonstrate scaling behavior equivalent to a $10^{13}$ token language corpus, up to the 1B parameters regime.

\begin{table}[H]
    \centering
    \caption{Power law constants (\textit{$\alpha$} in Equation~\ref{eq:one}) for different downstream tasks from Polaris benchmark when varying the number of parameters (Figure~\ref{fig:polaris_width_MPNN}).}  
    \resizebox{\textwidth}{!}{%
    \begin{tabular}{l|cccccccccccc}
    \toprule
    Method & T1 & T2 & T3 & T4 & T5 & T6 & T7 & T8 & T9 & T10 & T11 & T12 \\
    \midrule
    Probing	& 0.055	& 0.053	& 0.041	& 0.034	& 0.049	& 0.053	& 0.083	& 0.147	& 0.064 & 0.078 & 0.112 & 0.206 \\
    Finetuning	& 0.055	& 0.300	& 0.057	& 0.028	& 0.039	& 0.047	& 0.053	& 0.212	& 0.037	& 0.115 & 0.062 & 0.179 \\
    \bottomrule
    \end{tabular}
    }
    \label{tab:power-law:polaris}
\end{table}

\begin{table}[H]
    \centering
    \caption{Power law constants (\textit{$\beta$} in Equation~\ref{eq:two}) for pretraining different architectures when varying dataset sizes (Figure~\ref{fig:pretraining_one}).}
    \resizebox{\textwidth}{!}{%
    \begin{tabular}{l|ccccc}
    \toprule
    Model & Global Loss & L1000 (AVPR) & PCBA (AVPR) & PCQM4M\_G25 (R2) & PCQM4M\_N4 (R2) \\
    \midrule
    MPNN++ & 0.110 & 0.047 & 0.061 & 0.011 & 0.002 \\
    Transformer & 0.106 & 0.047 & 0.067 & 0.012 & 0.001 \\
    GPS++ & 0.106 & 0.047 & 0.067 & 0.012 & 0.001 \\
    \bottomrule
    \end{tabular}
    }
    \label{tab:power-law:data}
\end{table}